%% file: main.tex
\documentclass[11pt]{article}

\usepackage[final]{acl}

\usepackage{times}
\usepackage{latexsym}
\usepackage[T1]{fontenc}
\usepackage[utf8]{inputenc}
\usepackage{microtype}
\usepackage{inconsolata}
\usepackage{graphicx}

\usepackage{booktabs}
\usepackage{multirow}
\usepackage{amsmath}
\usepackage{amssymb}
\usepackage{xcolor}
\usepackage{subcaption}
\usepackage{xspace}
\usepackage{mwe} 
\usepackage[section]{placeins} 
\usepackage{arydshln}
\usepackage{dashrule}
\usepackage{pgfplots}
\usepackage[table]{xcolor}
\usepackage{xcolor}
\usepackage[most]{tcolorbox}

\usepackage{tabularx}
\usepackage{makecell}

\usepackage{xurl}

\usepgfplotslibrary{groupplots}
\pgfplotsset{compat=1.18}

\definecolor{lightgray}{gray}{0.9}

\newcommand{\sys}{\textsc{AdaPath}\xspace}

\newcommand{\cradd}[1]{#1}                    
\newenvironment{crblock}{}{}                

\title{\sys: Query-Adaptive Path-Finding via Path-Bank \\ for Multi-Hop Implicit Biomedical KGQA}

\author{
  \textbf{Jun Hyeong Kim\textsuperscript{1}},
  \textbf{Dongki Kim\textsuperscript{1}},
  \textbf{Yinhua Piao\textsuperscript{1,2}},
  \textbf{Sung Ju Hwang\textsuperscript{1,3}}
\\
\\
  \textsuperscript{1}KAIST,
  \textsuperscript{2}KAIST InnoCORE AI-CRED Institute,
  \textsuperscript{3}DeepAuto.ai
\\
  \texttt{\{tommykim0906, cleverki, yinhua.piao, sungju.hwang\}@kaist.ac.kr}
}

\begin{document}
\maketitle

\input{sections/00_abstract}


\input{sections/01_intro}            

\input{sections/02_related}            

\input{sections/03_method}            

\input{sections/04_experiments}            

\input{sections/05_analysis}            

\input{sections/06_conclusion}            

\section*{Limitations}
\input{sections/07_limitations}            

\section*{Acknowledgments}
This work was supported by Institute for Information \& communications
Technology Planning \& Evaluation (IITP) grant funded by the Korea
government (MSIT) (RS-2019-II190075, Artificial Intelligence Graduate
School Program (KAIST)), National Research Foundation of Korea (NRF)
grant funded by the Korea government (MSIT) (No. RS-2023-00256259), a
grant of the Korea Machine Learning Ledger Orchestration for Drug
Discovery Project (K-MELLODDY), funded by the Ministry of Health \&
Welfare and Ministry of Science and ICT, Republic of Korea (grant
number: RS-2024-00460870), the ``Advanced GPU Utilization Support
Program'' funded by the Government of the Republic of Korea (Ministry of
Science and ICT), the InnoCORE program of the Ministry of Science and
ICT (MSIT) (N10250153), Artificial intelligence industrial convergence
cluster development project funded by the Ministry of Science and ICT(MSIT, Korea) \&
Gwangju Metropolitan City, and the InnoCORE program of the Ministry of Science and ICT (N10260004).

\bibliography{anthology}

\clearpage
\appendix
\section*{Appendix}
\let\FloatBarrier\relax
\input{sections/A_appendix}

\end{document}

%% file: sections/00_abstract.tex
\begin{abstract}
Path-finding over knowledge graphs has become an effective way to ground LLM reasoning on multi-hop questions. However, biomedical QA introduces two distinct challenges that general-domain methods are not designed for: (i) queries do not expose intermediate reasoning and can be answered through multiple valid pathways, and (ii) biomedical knowledge graphs are densely connected, so path-finding methods easily take wrong turns.
To address these challenges, we propose \sys{}, a path-finding framework that retrieves query-adaptive meta-paths from \emph{Path-Bank}, which captures both query semantics and biomedical knowledge graph structure. \sys{} provides the missing cues in biomedical queries while effectively pruning dense knowledge graph neighborhoods during multi-hop reasoning.
We further release \textsc{BioStrat-QA}, a biomedical KGQA benchmark that stratifies multi-hop queries by how much intermediate reasoning they expose. 
Across biomedical KGQA benchmarks, \sys{} consistently outperforms baselines, sustaining meaningful path-finding even when multi-hop queries expose less surface information.
The source code is available at \cradd{\url{https://github.com/Jun-Hyeong-Kim/AdaPath}}.
\end{abstract}

%% file: sections/01_intro.tex
\section{Introduction}
\label{sec:intro}


\begin{figure}[!t]
  \centering
  \includegraphics[width=\columnwidth]{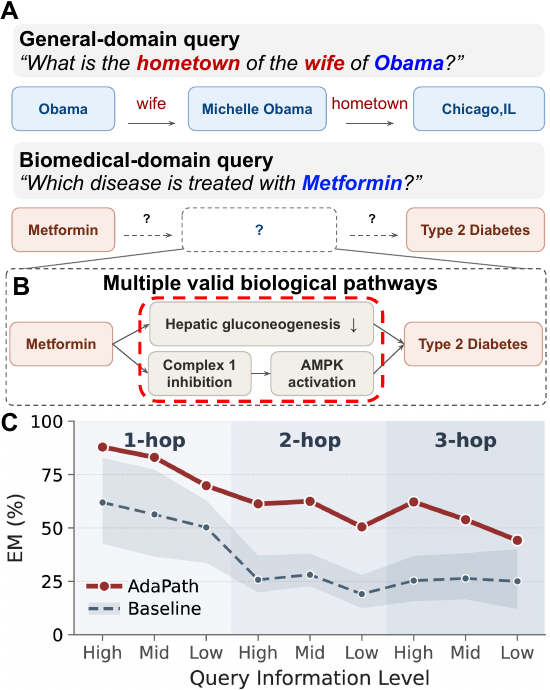}
  \caption{\textbf{Challenges of Biomedical KGQA and Motivation of \sys{}.}
  (A) The general-domain query includes traversal relations directly through ``\textit{wife}'' and ``\textit{hometown}'', constructing the intermediate path, while the biomedical query leaves such cues out. 
  (B) Several candidate intermediate paths exist for the biomedical query, as in the multiple mechanism chains from ``\textit{Metformin}'' to ``\textit{T2D}'', making it hard to identify the most relevant one.
  (C) \sys{} sustains performance relative to path-finding baselines on multi-hop biomedical queries with limited surface information.}
  \label{fig:intro_motivation}
\end{figure}

Recent large language models (LLMs) display increasingly sophisticated reasoning, and retrieval-augmented generation (RAG)~\citep{lewis2020retrieval} has become a widely
adopted approach for grounding LLM reasoning~\citep{wei2022chain,wangself} with external evidence
from knowledge sources. 
In the biomedical domain, question answering (QA) both demands domain-specific knowledge that LLMs often lack and requires chain of causalities rather than independent facts~\citep{barabasi2011network, menche2015uncovering, sung2021can, singhal2023large}.  
For instance, as shown in Figure~\ref{fig:intro_motivation}B, explaining \emph{why metformin treats type-2 diabetes} requires tracing a multi-step pathway from the drug through
protein-level interactions to its downstream effect on
gluconeogenesis~\citep{jin2021disease}.
Therefore, answering biomedical questions requires traversing multi-hop chains, necessitating the path-finding retrieval over biomedical knowledge graphs to ground LLM reasoning on the causal pathways~\citep{ogata1999kegg,milacic2024reactome} biomedical QA demands.

However, off-the-shelf path-finding methods~\citep{jiang2023structgpt,li2024chain,sun2024think,chen2024plan,ma2025think}
designed for general-domain Knowledge Graph Question Answering (KGQA)~\citep{wu2024stark, bollacker2008freebase, vrandevcic2014wikidata} struggle when applied to
biomedical KGQA due to its characteristics: 
\textbf{(i) Elusive guidance cues.}
Compared to general-domain queries, biomedical queries tend to directly state
a goal without exposing the intermediate steps needed to reach it,
lacking the salient cues for path-finding. Specifically, as shown in Figure~\ref{fig:intro_motivation}A, while general queries explicitly expose the cues for answering such as the relations and topic entities, cues for biomedical queries are hidden and non-trivial, facing a difficulty to guide multi-hop traversal. 
Compounding the difficulty, multiple valid biological pathways can typically lead to the same answer (Figure~\ref{fig:intro_motivation}B), making it more difficult to determine which cue is best suited to anchor traversal for any given query.
\textbf{(ii) Dense KG topology.} 
Existing path-finding methods~\citep{sun2024think,ma2025think} typically perform Breadth-First Search (BFS) over the KG from the topic entity toward the answer entity. However, the complex interconnectivity of biological pathways makes biomedical KGs far denser than general-domain ones (Table~\ref{tab:density}), forcing them to examine a large number of candidate entities. This complexity could mislead to follow high-degree biological hubs or weakly relevant connections, leading to spurious paths that deviate from the biologically relevant reasoning chain.

To address these issues, we propose \textbf{\sys{}}, a path-finding framework guided by query-adaptive meta-paths, explicitly supplying the intermediate cues that biomedical queries often leave implicit while enhancing the dense KG traversal via pruning.
We first construct a \textbf{Path-Bank}, a repository of typed meta-paths mined from training queries. 
These meta-paths encode reusable intermediate cues that are often missing from biomedical queries, while reflecting both query semantics and the structural connectivity of the biomedical KG.
At inference time, \sys{} adaptively retrieves query-relevant meta-paths from Path-Bank and uses them as pruning guidance for candidate expansion in the dense biomedical KG. 
This enables to recover intermediate cues absent from the query while focusing traversal on biologically plausible directions.
Consequently, \sys{} avoids the indiscriminate expansion of BFS while preserving the flexibility to follow multiple valid reasoning pathways toward an answer (Figure~\ref{fig:intro_motivation}C).

\begin{table}[!tb]
  \centering
  \small
  \setlength{\tabcolsep}{3pt}
  \renewcommand{\arraystretch}{1.05}
  \begin{tabular*}{\columnwidth}{@{\extracolsep{\fill}}llrrr@{}}
    \toprule
    \textbf{Domain} & \textbf{KG}
      & \textbf{Node \#} & \textbf{Edge \#}
      & \textbf{Degree} \\
    \midrule
    \multirow{2}{*}{\textit{General}}
      & Amazon & 1.04\,M &  9.44\,M &  18.2 \\
      & MAG    & 1.87\,M & 39.80\,M &  43.5 \\
    \midrule
    \textit{Biomedicine}
      & \textbf{Prime} & \textbf{129\,K} & \textbf{8.10\,M} & \textbf{125.2} \\
    \bottomrule
  \end{tabular*}
  \caption{\textbf{Statistics on Knowledge Graphs across different domains.} \cradd{Degree is the average node degree $2|E|/|V|$, with each edge treated as undirected.}}
  \label{tab:density}
\end{table}

\cradd{On the other hand,} evaluating biomedical path-finding requires controlled tests of whether a method can recover missing intermediate cues and remain reliable as reasoning depth increases. 
However, existing biomedical KGQA evaluations do not explicitly disentangle hop depth from the amount of intermediate reasoning exposed in the query. 
We therefore release \textsc{BioStrat-QA}, a biomedical KGQA benchmark that stratifies multi-hop queries by surface exposure of intermediate reasoning. 
It derives queries from shared reasoning chains and organizes them into \emph{explicit}, \emph{implicit}, and \emph{bare} levels, enabling systematic measurement of path-finding robustness as surface cues decrease and reasoning paths extend up to three hops.

We experimentally validate \sys{} on diverse biomedical KGQAs including \textsc{BioStrat-QA}, STaRK-Prime, and MedDDx, and we observe that \sys{} consistently outperforms baselines, including path-finding methods, across all three benchmarks, particularly on multi-hop queries with limited intermediate cues.
Path-level analyses further show that \sys{} retrieves paths that both align with ground-truth reasoning chains and surface semantically related alternatives, matching the dual challenge of implicit cues and multiple valid pathways in biomedical KGQA.
We summarize our contribution as follows:
\vspace{-0.1in}
\begin{itemize}\setlength\itemsep{2pt}
    \item We introduce \sys{}, a path-finding framework where meta-paths from a \emph{Path-Bank} reflect both query semantics and biomedical KG structure, supplying implicit cues and filtering dense KG \cradd{neighborhoods}.
    \item We release \textsc{BioStrat-QA}, a biomedical KGQA benchmark designed to systematically measure path-finding ability under these challenges, stratifying queries along \emph{explicit} / \emph{implicit} / \emph{bare} levels of intermediate-information exposure and hop depths up to three.
    \item \sys{} consistently outperforms baselines on biomedical KGQA, with gains over path-finding baselines most pronounced on multi-hop, surface-implicit queries where dense KG topology most easily misleads path-finding.
\end{itemize}



\FloatBarrier 

%% file: sections/02_related.tex
\section{Related Work}
\label{sec:related}

\subsection{LLM Reasoning for Knowledge Graph Question Answering (KGQA)}

Retrieval-augmented generation~\citep{lewis2020retrieval} grounds LLM reasoning~\citep{wangself,wei2022chain} by injecting external knowledge into the prompt, mitigating hallucination and stale parametric knowledge.
For KGQA, fact-retrieval methods~\citep{pan2024unifying} typically extract triples or short paths matching the query and inject them into the LLM prompt as evidence. 
Chain-of-Knowledge~\citep{li2024chain} retrieves evidence chains across heterogeneous sources such as KGs, Wikipedia, and tables, injecting them into the prompt without exploiting explicit graph structure. KGARevion~\citep{su2025kgarevion} targets biomedical QA by having the LLM generate candidate triples, verifying them against the KG, and revising the chain when verification fails.

\subsection{Path-Finding Methods for KGQA}

Path-finding methods construct multi-hop reasoning chains explicitly by traversing the KG hop by hop, providing the LLM with structured evidence paths rather than the local triples fact-retrieval methods return. 
StructGPT~\citep{jiang2023structgpt} treats the KG as a structured store accessed through specialized query interfaces, expanding the single-hop neighborhood at each step and having the LLM filter the result. 
Think-on-Graph (ToG)~\citep{sun2024think} performs LLM-guided beam search over the KG, maintaining top-$N$ reasoning paths at each hop. 
Think-on-Graph 2.0 (ToG-2)~\citep{ma2025think} hybridizes beam search with text-based retrieval and iterative refinement so that graph and context mutually refine each other. 
Plan-on-Graph (PoG)~\citep{chen2024plan} decomposes the question into sub-objectives and self-corrects exploration via guidance, memory, and reflection. 
These methods are designed for general-domain KGs, where per-hop neighborhoods are small and queries tend to surface their reasoning chains. However, they struggle to transfer to biomedical KGQA.

\subsection{Biomedical KGQA Benchmarks}

Biomedical KGQA evaluates whether retrieval-augmented systems can navigate biomedical relations on a knowledge graph and reach the correct answer entity, commonly used to benchmark domain reasoning in clinical, pharmacological, and disease-mechanism settings. 
STaRK-Prime~\citep{wu2024stark} provides a template-driven retrieval benchmark built over PrimeKG~\citep{chandak2023building}, with questions mostly at shallow hop depths and reasonably explicit phrasing. 
MedDDx~\citep{su2025kgarevion} builds a differential-diagnosis variant on STaRK-Prime by adding semantically similar entities as distractors, ordered from Basic through Intermediate to Expert as distractor similarity to the gold answer increases.

%% file: sections/03_method.tex
\section{Method}
\label{sec:method}

\begin{figure*}[!t]
  \centering
  \includegraphics[width=1\textwidth]{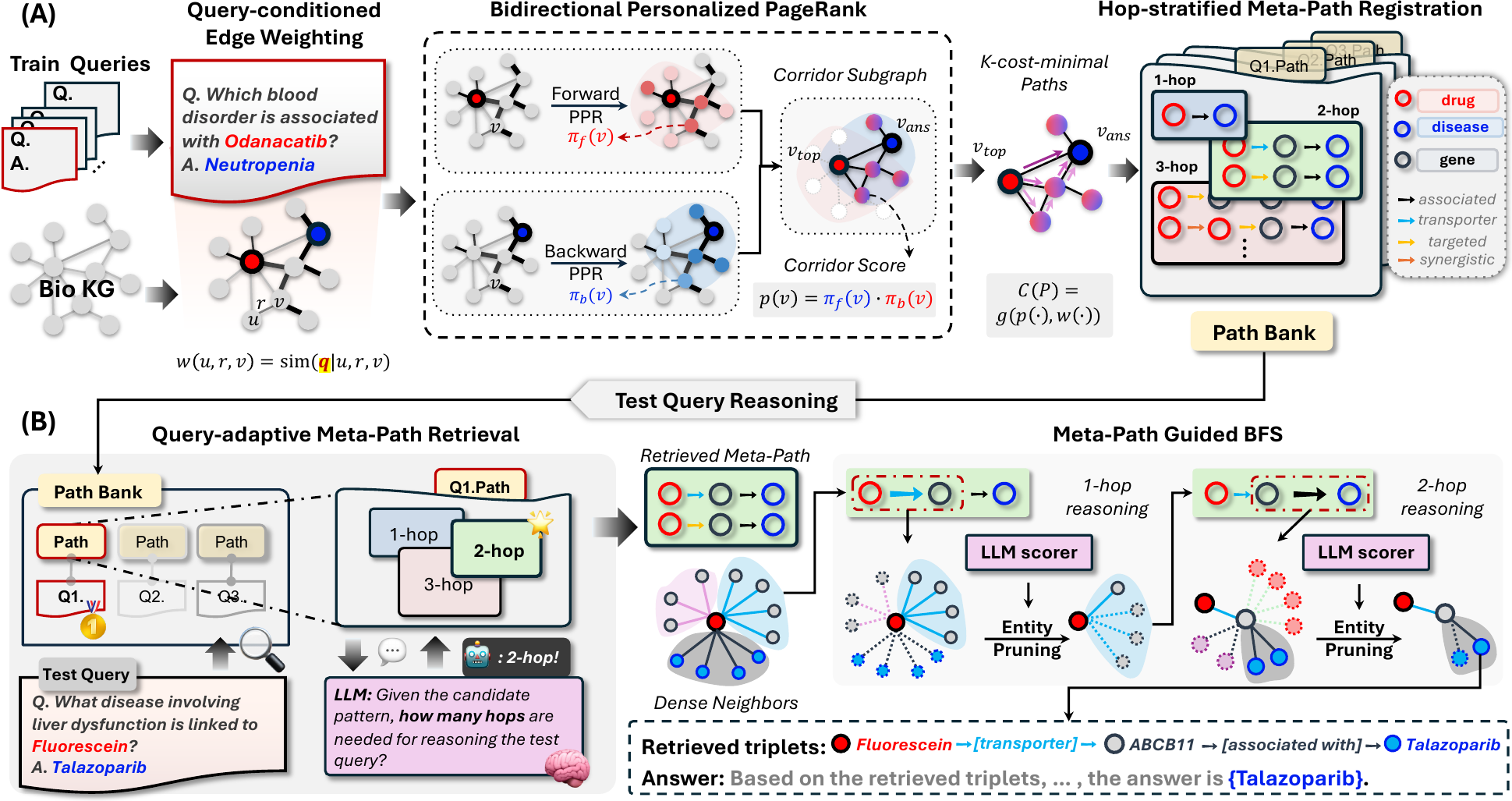}
  \caption{\textbf{\sys{} Overview.}
  \textbf{(A) Path-Bank Construction} (Section~\ref{sec:method:pathbank}). We construct a \textbf{Path-Bank}, a repository of meta-paths mined from training queries. These meta-paths encode reusable intermediate cues and reflect both query semantics and the structural connectivity of the biomedical KG.
  \textbf{(B) Query-Adaptive Meta-Path Guided Path-Finding} (Section~\ref{sec:method:pipeline}). At inference, \sys{} adaptively retrieves test-query-relevant meta-paths from Path-Bank and uses them as pruning guidance for candidate expansion in the dense biomedical KG.}
  \label{fig:main_framework}
\end{figure*}

\sys{} is a two-stage path-finding framework guided by query-adaptive meta-paths retrieved from a Path-Bank (Figure~\ref{fig:main_framework}). 
In Section~\ref{sec:method:pathbank}, we construct Path-Bank by mining meta-paths from training queries, which reflect both query semantics and biomedical KG structure. 
In Section~\ref{sec:method:pipeline}, at the inference stage, \sys{} adaptively retrieves query-relevant meta-paths from the constructed Path-Bank and uses them as pruning guidance for candidate expansion in the dense biomedical KG. 
Alongside the framework, in Section~\ref{sec:method:bench} we construct \textsc{BioStrat-QA}, a biomedical KGQA benchmark for systematic evaluation of path-finding methods, with multi-hop queries derived from shared reasoning chains and stratified by surface exposure of intermediate reasoning.

\subsection{Path-Bank Construction}
\label{sec:method:pathbank}

To address the challenges (implicit-query and dense-KG) mentioned in Section~\ref{sec:intro}, query-adaptive guidance is necessary for traversal at inference. \sys{} provides this guidance as meta-paths retrieved from Path-Bank, mined offline from training queries. To construct a rich and reliable pool of meta-paths from the dense biomedical KG, Path-Bank construction proceeds in three stages: \textbf{(i)} Query-conditioned edge weighting, \textbf{(ii)} Bidirectional Personalized PageRank (PPR), and \textbf{(iii)} Hop-stratified meta-path registration.

\paragraph{Query-conditioned Edge Weighting.}
Path-Bank construction uses Personalized PageRank (PPR)~\citep{jeh2003scaling}, a random walk over the KG that estimates each node's structural proximity to designated seeds. 
To mine paths that reflect query semantics in addition to KG structure, we weight each edge by how strongly its endpoints and relation align with the train query $q$, combining the three similarities into a single weight $w_{u,r,v}$ for the edge connecting $u$ to $v$ through relation $r$:
\begin{equation}
\begin{aligned}
w_{u,r,v} = &\; \beta\, \mathrm{sim}(q, r) \\
          &+ \tfrac{1-\beta}{2}\bigl[\mathrm{sim}(q, u) + \mathrm{sim}(q, v)\bigr].
\end{aligned}
\label{eq:edge_weight}
\end{equation}
$\mathrm{sim}$ is a per-query normalized BM25~\citep{lin2022pretrained} similarity to biomedical KG's node and relation descriptions, and $\beta \in [0,1]$ trades off relation against endpoint terms.


\paragraph{Bidirectional Personalized PageRank.}
To select nodes that can form routes between topic and answer entity for each training query, we run forward and backward PPR walks seeded at the two entities, both biased by the query-conditioned edge weights $w_{u,r,v}$. 
Let $M$ be the transition matrix obtained by normalizing $w_{u,r,v}$ over each node's outgoing edges, $\eta$ the teleport probability, and $s_{\text{top}}, s_{\text{ans}}$ the one-hot seed indicators at the topic and answer entities. 
The forward and backward stationary distributions $\pi_f$ and $\pi_b$ satisfy
\begin{align}
\pi_f &= (1-\eta)\, M\, \pi_f + \eta\, s_{\text{top}},
\label{eq:ppr_fwd} \\
\pi_b &= (1-\eta)\, M\, \pi_b + \eta\, s_{\text{ans}}.
\label{eq:ppr_bwd}
\end{align}
The element-wise product of $\pi_f$ and $\pi_b$ yields a \textit{corridor score}
\begin{equation}
\rho(v) = \pi_f(v) \cdot \pi_b(v).
\label{eq:corridor}
\end{equation}
The top-$K$ nodes by corridor score form the query-specific \textit{corridor subgraph}, semantically aligned with the query and structurally central to both topic and answer entities.

\paragraph{Hop-stratified Path-Bank Registration.}
To enrich each training query's Path-Bank entry with meta-paths across multiple hops, we extract the top-$n$ cost-minimizing paths from the topic entity $v_{\text{top}}$ to the answer entity $v_{\text{ans}}$ within the corridor subgraph using Yen's $k$-shortest-paths algorithm~\citep{yen1971finding}. 
The path cost
\begin{equation}
  c(P) = \sum_{(u,r,v) \in P} \bigl[\alpha\, w_{u,r,v} + (1-\alpha)\, \bar{\rho}(u,v)\bigr]^{-1},
\label{eq:yen_cost}
\end{equation}
decreases in both the query-conditioned edge weight $w_{u,r,v}$ from Eq.~\ref{eq:edge_weight} and the average endpoint corridor score $\bar{\rho}(u,v) = \tfrac{1}{2}(\rho(u) + \rho(v))$, with $\alpha \in [0,1]$ trading off the two. 
We convert each extracted path to its meta-path and bin it by hop length as the Path-Bank entry for $q$.

\subsection{Query-Adaptive Meta-Path Guided Path-Finding}
\label{sec:method:pipeline}

For each test query $q_{\text{test}}$, it retrieves from Path-Bank a query-adaptive set of meta-paths and uses them to guide BFS over the biomedical knowledge graph (bioKG) from $q_{\text{test}}$'s topic entity. 
The retrieved meta-paths, drawn from training queries similar to $q_{\text{test}}$, supply the intermediate cues $q_{\text{test}}$ leaves implicit on its surface. 
Restricting BFS expansion to the matched relation-type transitions filters the dense biomedical KG to directions suitable for $q_{\text{test}}$.

\paragraph{Path-Bank Retrieval.}
We retrieve $q_{\text{test}}$-adaptive meta-paths from Path-Bank in two steps: candidate query ranking, then meta-path filtering. 
With $q_{\text{test}}$'s topic entity linked to a bioKG node, we restrict candidates to training queries with the same topic entity type and rank each $q_i$ by
\begin{equation}
\mathrm{sim}(q, q_i) = \lambda\, \widetilde{s}_{\text{SBERT}}(q, q_i)
                  + (1-\lambda)\, \widetilde{s}_{\text{BM25}}(q, q_i),
\label{eq:hybrid_sim}
\end{equation}
where $\widetilde{s}_{\text{SBERT}}$ and $\widetilde{s}_{\text{BM25}}$ are per-test-query min-max normalized sentence-embedding~\citep{reimers2019sentence} and BM25~\citep{lin2022pretrained} similarities, and $\lambda \in [0,1]$ trades off the two. 
From the meta-paths of the top-$k$ ranked training queries, we retain those consistent with an LLM-inferred answer type and hop length, and realizable as walks from $v_{\text{top}}$ on bioKG.

\begin{figure}[t]
\centering
\includegraphics[width=\columnwidth]{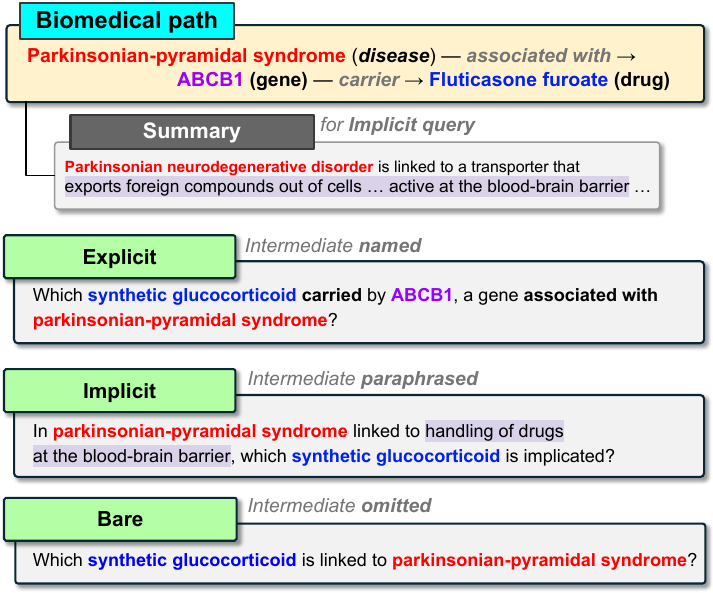}
\caption{\textbf{\textsc{BioStrat-QA} Query Stratification.}
Three queries are generated from one reference path, differing in how
much intermediate information reaches the surface: named
(\emph{Explicit}), summarized without naming (\emph{Implicit}), or
omitted (\emph{Bare}).}
\label{fig:biotog_example}
\end{figure}

\paragraph{Meta-Path Guided BFS.}
Guided by the retrieved meta-paths, \sys{} performs BFS over bioKG from the topic entity $v_{\text{top}}$, retrieving candidate paths for $q_{\text{test}}$. 
At each hop, \sys{} keeps only neighbors whose incoming relation and node type match a retrieved meta-path, mitigating the wrong-turn risk BFS faces at densely connected hubs. 
The surviving candidates are further narrowed by sentence-BERT similarity to $q_{\text{test}}$, and the next-hop frontier is selected by LLM scoring under a depth-specific sub-query. 
When the final hop is reached, the assembled paths are passed to the LLM with $q_{\text{test}}$ to ground its final answer.

\subsection{\textsc{BioStrat-QA} Construction}
\label{sec:method:bench}

We also construct \textsc{BioStrat-QA} to systematically evaluate path-finding when biomedical queries leave intermediate reasoning implicit, stratifying multi-hop queries by how much of it surfaces.
Built atop PrimeKG~\citep{chandak2023building}, \textsc{BioStrat-QA} extends STaRK-Prime~\citep{wu2024stark}'s 2-hop query templates with additional single-hop and 3-hop templates, retaining only those with sufficient grounding in PrimeKG. 
Full templates and synthesis prompts appear in Appendix~\ref{sec:appendix:dataset}.

\paragraph{Stratification by intermediate-information level.}
For each reference path, we synthesize three parallel queries using GPT-5.4~\citep{openai2026gpt54} at distinct levels of intermediate-information exposure: \textbf{Explicit}, \textbf{Implicit}, and \textbf{Bare}. 
All three share the path's topic and answer entities, differing only in surface exposure of intermediate reasoning. 
\textbf{Explicit} names every intermediate entity and relation along the path. 
\textbf{Implicit} replaces intermediate entities with paraphrased descriptions preserving the underlying mechanism, using a pathway summary for multi-hop paths. 
\textbf{Bare} retains only the topic entity and the answer type, stripping all intermediate cues. 
For 1-hop paths with no intermediate node, the three levels collapse to differences in how directly the answer-bearing relation is named, with \emph{Bare} omitting the relation entirely. 
Figure~\ref{fig:biotog_example} shows a 2-hop example on a \textit{disease $\to$ gene $\to$ drug} path.




%% file: sections/04_experiments.tex
\section{Experiments}
\label{sec:experiments}


\subsection{Datasets}
\label{sec:experiments:datasets}

\begin{table*}[!t]
  \centering\footnotesize
  \setlength{\tabcolsep}{4pt} 
  \setlength{\dashlinedash}{0.5pt}
  \setlength{\dashlinegap}{1.5pt}
  \renewcommand{\arraystretch}{1.05}
  \resizebox{\textwidth}{!}{ 
  \begin{tabular}{@{}ll cccc cccc cccc c@{}}
    \toprule
    \multirow{2}{*}{\textbf{Category}} & \multirow{2}{*}{\textbf{Method}} & \multicolumn{4}{c}{\textbf{Explicit}} & \multicolumn{4}{c}{\textbf{Implicit}} & \multicolumn{4}{c}{\textbf{Bare}} & \multirow{2}{*}{\textbf{Overall}} \\
    \cmidrule(lr){3-6} \cmidrule(lr){7-10} \cmidrule(lr){11-14}
    & & 1-hop & 2-hop & 3-hop & Avg & 1-hop & 2-hop & 3-hop & Avg & 1-hop & 2-hop & 3-hop & Avg & \\
    \midrule
    \rowcolor{lightgray}
    \multicolumn{15}{c}{\texttt{Llama-3.1-70B}} \\
    \midrule
    \multirow{3}{*}{\textbf{LLM only}}
    & IO                                                 & 45.8 & 35.8 & 34.1 & 39.2 & 41.6 & 39.4 & 35.5 & 39.5 & 42.3 & 33.3 & 33.6 & 36.7 & 38.5 \\
    & CoT                                                & 46.2 & 36.4 & 32.7 & 39.4 & 40.0 & \underline{40.0} & 32.7 & 38.7 & 41.4 & 32.8 & 31.8 & 35.8 & 37.9 \\
    & SC                                                 & 43.5 & 35.6 & 33.6 & 38.2 & 38.7 & 37.9 & 37.8 & 38.2 & 43.0 & 32.6 & 30.9 & 36.1 & 37.5 \\
    
    \noalign{\vskip 2pt}    \cdashline{1-15}    \noalign{\vskip 2pt}
    
    \multirow{2}{*}{\textbf{Fact-retrieval}}
    & CoK                                                & 49.2 & \underline{38.5} & 31.8 & \underline{41.2} & 44.2 & 39.8 & \underline{40.6} & 41.6 & 45.3 & \underline{36.4} & 38.2 & \underline{40.0} & \underline{41.0} \\
    & KGARevion                                          & 30.7 & 27.6 & 34.1 & 30.0 & 30.7 & 31.2 & 36.4 & 32.0 & 30.4 & 21.7 & 33.2 & 27.0 & 29.7 \\

    \noalign{\vskip 2pt}    \cdashline{1-15}    \noalign{\vskip 2pt}
    
    \multirow{5}{*}{\textbf{Path-finding}}
    & StructGPT                                          & 42.6 & 21.1 & \underline{36.9} & 32.0 & 36.6 & 26.3 & 33.6 & 31.5 & 33.6 & 17.5 & \underline{40.1} & 27.6 & 30.4 \\
    & ToG                                                & 55.8 & 37.1 & 32.7 & 43.2 & 49.4 & 38.1 & 38.2 & 42.3 & 43.7 & 28.0 & 31.8 & 34.5 & 40.0 \\
    & ToG-2                                              & 66.6 & 19.8 & 16.1 & 36.5 & 62.0 & 22.7 & 16.6 & 36.1 & 60.9 & 18.3 & 12.0 & 32.9 & 35.2 \\
    & PoG                                                & \underline{82.8} & 25.0 & 15.7 & 44.7 & \underline{77.3} & 25.3 & 17.1 & \underline{43.1} & \underline{62.9} & 12.4 & 16.1 & 31.8 & 39.9 \\
    \rowcolor[HTML]{F4EBFF}
    \cellcolor{white} & \textbf{AdaPath}                    & \textbf{87.9} & \textbf{61.3} & \textbf{62.2} & \textbf{71.3} & \textbf{83.1} & \textbf{62.5} & \textbf{53.9} & \textbf{68.5} & \textbf{69.8} & \textbf{50.5} & \textbf{44.2} & \textbf{56.5} & \textbf{65.5} \\
    \midrule
    \rowcolor{lightgray}
    \multicolumn{15}{c}{\texttt{Qwen-2.5-72B}} \\
    \midrule
    \multirow{3}{*}{\textbf{LLM only}}
    & IO                                                 & 41.6 & 29.5 & 27.6 & 33.6 & 39.4 & 34.3 & 30.9 & 35.6 & 43.0 & \underline{32.2} & 35.5 & 36.8 & 35.3 \\
    & CoT                                                & 39.4 & 31.6 & 30.4 & 34.3 & 35.5 & 35.4 & 36.4 & 35.6 & 37.8 & 30.5 & \underline{37.8} & 34.6 & 34.8 \\
    & SC                                                 & 41.0 & 32.6 & 32.7 & 35.7 & 34.3 & 36.2 & \underline{38.2} & 35.9 & 38.9 & 30.3 & 36.4 & 34.6 & 35.4 \\

    \noalign{\vskip 2pt}    \cdashline{1-15}    \noalign{\vskip 2pt}
    
    \multirow{2}{*}{\textbf{Fact-retrieval}}
    & CoK                                                & 47.6 & \underline{33.9} & 33.2 & 38.9 & 40.0 & \underline{36.8} & 35.0 & 37.7 & 44.4 & 30.5 & \underline{37.8} & \underline{37.0} & 37.8 \\
    & KGARevion                                          & 25.4 & 18.7 & 33.2 & 23.9 & 23.3 & 22.7 & 35.0 & 25.2 & 21.7 & 17.9 & 27.6 & 21.1 & 23.4 \\

    \noalign{\vskip 2pt}    \cdashline{1-15}    \noalign{\vskip 2pt}
    
    \multirow{5}{*}{\textbf{Path-finding}}
    & StructGPT                                          & 58.1 & 18.9 & \underline{34.1} & 36.2 & 52.6 & 20.6 & 35.5 & 35.2 & 36.6 & 9.7  & 22.6 & 22.0 & 31.2 \\
    & ToG                                                & 55.6 & 31.4 & 30.4 & 40.2 & 51.9 & 34.7 & 34.6 & 41.1 & 48.3 & 27.4 & 32.3 & 36.1 & 39.1 \\
    & ToG-2                                              & 70.9 & 27.8 & 30.0 & 44.2 & 65.9 & 27.4 & 17.5 & 39.9 & 58.8 & 22.5 & 22.1 & 35.9 & 40.0 \\
    & PoG                                                & \underline{76.0} & 29.9 & 24.9 & \underline{46.1} & \underline{72.1} & 32.8 & 30.4 & \underline{46.9} & \underline{59.7} & 9.7  & 10.1 & 28.3 & \underline{40.4} \\
    \rowcolor[HTML]{F4EBFF}
    \cellcolor{white} & \textbf{AdaPath}                    & \textbf{84.9} & \textbf{58.3} & \textbf{53.9} & \textbf{67.4} & \textbf{83.3} & \textbf{59.2} & \textbf{44.7} & \textbf{65.5} & \textbf{71.9} & \textbf{44.8} & \textbf{41.0} & \textbf{54.1} & \textbf{62.3} \\
    \bottomrule
  \end{tabular}
  }
  \caption{\textbf{Main results on \textsc{BioStrat-QA}.}
  Exact-match (\%) per (information-level, hop-depth) cell;
  \textbf{Overall} averages all nine cells.
  Per-column best is \textbf{bold} and second-best is \underline{underlined}
  (ties share rank). Small-backbone (Llama-3.1-8B, Qwen-2.5-7B) results are in
  Appendix~\ref{sec:appendix:extended}.}
  \label{tab:main_results_biotog}
\end{table*}

We evaluate on three biomedical KGQA datasets. \textbf{\textsc{BioStrat-QA}}, introduced in
Section~\ref{sec:method:bench}, stratifies multi-hop queries into explicit,
implicit, and bare information levels, enabling fine-grained evaluation
of path-finding across query difficulty.
\textbf{STaRK-Prime}~\citep{wu2024stark} contributes two splits: a
\textit{Synthesized} split of 2{,}801 LLM-synthesised template-based queries, and
a \textit{Human-generated} split of 98 naturally phrased, human-written queries
with manually verified answers. 
\textbf{MedDDx}~\citep{su2025kgarevion} contributes 1{,}769 multi-choice differential-diagnosis queries built from STaRK-Prime by adding semantically similar distractors, with \textit{Basic}, \textit{Intermediate}, and \textit{Expert} bins ordered by increasing distractor similarity to the gold answer.
For STaRK-Prime and MedDDx, we extract topic entities from each query during preprocessing to serve as starting nodes for path-finding methods (Appendix~\ref{sec:appendix:topic_extraction}).


\subsection{Baselines}
\label{sec:experiments:baselines}
We compare against three baseline categories. 
\textbf{LLM-only baselines} establish a floor without any KG, including IO (direct prompting), Chain-of-Thought (CoT)~\citep{wei2022chain}, and Self-Consistency (SC)~\citep{wangself}. 
\textbf{Fact-level retrieval methods} consult the KG without traversing it, including CoK~\citep{li2024chain} and KGARevion~\citep{su2025kgarevion}.
\textbf{Path-finding methods}, our most direct comparison, traverse the KG hop by hop. We compare with StructGPT~\citep{jiang2023structgpt}, 
Plan-on-Graph (PoG)~\citep{chen2024plan}, Think-on-Graph (ToG)~\citep{sun2024think}, and Think-on-Graph 2.0 (ToG-2)~\citep{ma2025think}. For ToG-2, we used semi-structured KG node descriptions as the per-node text source.


\subsection{Experimental Settings}
\label{sec:experiments:impl}
We evaluate \sys{} and all baselines using two families of open-weight LLMs, Llama-3.1-Instruct~\citep{grattafiori2024llama} and Qwen-2.5-Instruct~\citep{yang2024qwen25}. 
We use the bioKG from STaRK-Prime~\citep{wu2024stark}, PrimeKG augmented with per-node textual descriptions.
\cradd{Path-finding baselines share the same topic-entity links, bioKG, backbone, and three-hop search limit.}
We report exact match (EM) on \textsc{BioStrat-QA} and STaRK-Prime, and accuracy over four candidate answers for each multiple-choice query on MedDDx.
\cradd{Metric details and \sys{} hyperparameters are in Appendices~\ref{sec:appendix:metrics} and~\ref{sec:appendix:hyperparams}.}



\begin{table}[!t]
  \centering\footnotesize
  \setlength{\tabcolsep}{3pt}
  \setlength{\dashlinedash}{0.5pt}
  \setlength{\dashlinegap}{1.5pt}
  \renewcommand{\arraystretch}{1.05}
  \resizebox{\columnwidth}{!}{
  \begin{tabular}{@{}l cc cccc@{}}
    \toprule
    \multirow{2}{*}{\textbf{Method}} & \multicolumn{2}{c}{\textbf{STaRK-Prime}} & \multicolumn{4}{c}{\textbf{MedDDx}} \\
    \cmidrule(lr){2-3} \cmidrule(lr){4-7}
    & Synthesized & \begin{tabular}{@{}c@{}}Human-\\generated\end{tabular} & Basic & Inter. & Expert & All \\
    \midrule
    \rowcolor{lightgray}
    \multicolumn{7}{c}{\texttt{Llama-3.1-70B}} \\
    \midrule
    IO                                                 & 18.6 & 20.4 & 44.5 & 39.2 & 41.6 & 40.6 \\
    CoT                                                & 16.6 & 21.4 & 50.2 & 40.3 & 40.0 & 41.6 \\
    SC                                                 & 16.7 & 21.4 & 56.3 & 42.5 & 45.3 & 45.2 \\
    
    \noalign{\vskip 2pt}    \cdashline{1-7}    \noalign{\vskip 2pt}
    
    CoK                                                & 18.2 & 17.3 & 51.4 & 44.1 & 46.8 & 45.8 \\
    KGARevion                                          & 18.9 & 18.4 & 56.3 & 44.1 & 47.4 & 46.7 \\

    \noalign{\vskip 2pt}    \cdashline{1-7}    \noalign{\vskip 2pt}
    
    StructGPT                                          & 37.6 & 30.6 & 56.3 & \underline{48.3} & 45.1 & 48.6 \\
    ToG                                                & 32.1 & 22.4 & \underline{60.4} & \underline{48.3} & \underline{47.6} & \underline{49.8} \\
    ToG-2                                              & 33.5 & 34.7 & 37.1 & 34.4 & 30.9 & 33.8 \\
    PoG                                                & \underline{41.4} & \underline{40.8} & 55.5 & 47.0 & 43.9 & 47.3 \\
    \rowcolor[HTML]{F4EBFF} \textbf{AdaPath}             & \textbf{46.9} & \textbf{43.9} & \textbf{65.3} & \textbf{55.1} & \textbf{52.4} & \textbf{55.8} \\
    \midrule
    \rowcolor{lightgray}
    \multicolumn{7}{c}{\texttt{Qwen-2.5-72B}} \\
    \midrule
    IO                                                 & 16.1 & 17.3 & 40.4 & 39.4 & 35.4 & 38.4 \\
    CoT                                                & 14.1 & 16.3 & 50.2 & 40.2 & 38.7 & 41.2 \\
    SC                                                 & 15.0 & 18.4 & 50.2 & 41.0 & 40.0 & 42.0 \\

    \noalign{\vskip 2pt}    \cdashline{1-7}    \noalign{\vskip 2pt}
    
    CoK                                                & 14.9 & 15.3 & 44.9 & 41.6 & 39.5 & 41.5 \\
    KGARevion                                          & 16.0 & 15.3 & 53.5 & 41.9 & 39.3 & 42.8 \\

    \noalign{\vskip 2pt}    \cdashline{1-7}    \noalign{\vskip 2pt}
    
    StructGPT                                          & 32.6 & 30.6 & \underline{58.8} & \underline{50.0} & 42.4 & \underline{49.2} \\
    ToG                                                & 32.0 & 25.5 & 54.3 & 48.8 & \underline{43.5} & 48.1 \\
    ToG-2                                              & 31.9 & \underline{35.7} & 39.6 & 36.1 & 32.3 & 35.6 \\
    PoG                                                & \underline{33.5} & 30.6 & 44.9 & 42.7 & 41.0 & 42.6 \\
    \rowcolor[HTML]{F4EBFF} \textbf{AdaPath}             & \textbf{46.7} & \textbf{39.8} & \textbf{62.0} & \textbf{54.5} & \textbf{51.8} & \textbf{54.8} \\
    \bottomrule
  \end{tabular}
  }
  \caption{\textbf{Results on STaRK-Prime and MedDDx.}
  Performance comparison across different methods. 
  Best results are \textbf{bolded}, and second-best results are \underline{underlined}.}
  \label{tab:stark_medddx_results}
\end{table}

\subsection{Results}
\label{sec:experiments:main}
As shown in Table~\ref{tab:main_results_biotog}, we evaluate \sys{} and the baselines on \textsc{BioStrat-QA} across query information levels and hop depths. 
\sys{} consistently outperforms all retrieval and path-finding baselines across every information level and hop depth. 
Path-finding baselines, in particular, degrade sharply on multi-hop queries with low surface information. 
Lacking surface cues and confronting the dense biomedical KG, they accumulate wrong turns hop by hop, with several multi-hop \textit{bare} cells even falling below the LLM-only floor. 
Even on \textit{implicit} and \textit{bare} queries where surface cues are scarce or absent, \sys{} sustains robust and consistently high performance through Path-Bank meta-paths mined per-query from both query semantics and KG structure.
\sys{} also outperforms all baselines on the external datasets STaRK-Prime and MedDDx (Table~\ref{tab:stark_medddx_results}), generalizing across synthesized and  human-generated queries on STaRK-Prime and maintaining robustness across difficulty levels on MedDDx.
\cradd{We further evaluate \sys{} on the smaller Llama-3.1-8B and
Qwen-2.5-7B backbones, where it retains the same trend and leads all
baselines (Appendix~\ref{sec:appendix:extended}).}

\FloatBarrier

%% file: sections/05_analysis.tex
\subsection{Path-level Analysis of \sys{}}
\label{sec:analysis:pathrecover}

To further analyze whether retrieved paths are quantitatively and qualitatively related to solving biomedical queries, we assess whether the paths retrieved during traversal support answering the query (Table~\ref{tab:recall_ctx_relevance}). 
For each test query and at each reasoning depth, \textit{Recall} measures how much of the reference biomedical path is recovered in the retrieved set through exact triplet match. 
Additionally, given that multiple valid biological pathways typically lead to the same answer for any given biomedical query, we further measure \textit{Context Relevance (CR)}~\citep{xiang2026use}.
Scored by GPT-4o-mini~\citep{openai2024gpt4omini} as an LLM judge, \textit{Context Relevance} assesses whether the retrieved paths as a whole form a biomedical mechanism semantically aligned with both the query and the reference path, capturing whether the path-finding genuinely contributes to solving the query. 
\sys{} outperforms other path-finding methods on both \textit{Recall} and \textit{Context Relevance} across all conditions, including queries with limited surface cues.
The retrieved meta-paths from Path-Bank, semantically and structurally aligned with the test query, lead \sys{} to retrieve paths that remain semantically connected to the query's reasoning. 
\cradd{These meta-paths also remain traversable from the topic entity to
the answer at a meaningful rate even after exact ground-truth matches are
excluded, surfacing valid alternative schemas rather than a single
reference path (Appendix~\ref{sec:appendix:metapath}).}

\begin{table*}[!t]
  \centering\small
  \setlength{\tabcolsep}{4pt}
  \renewcommand{\arraystretch}{1.05}
  \begin{tabularx}{\textwidth}{@{}l *{12}{>{\centering\arraybackslash}X}@{}}
    \toprule
    & \multicolumn{6}{c}{\textbf{Llama-3.1-70B}} & \multicolumn{6}{c}{\textbf{Qwen-2.5-72B}} \\
    \cmidrule(lr){2-7} \cmidrule(lr){8-13}
    & \multicolumn{2}{c}{\textbf{Explicit}} & \multicolumn{2}{c}{\textbf{Implicit}} & \multicolumn{2}{c}{\textbf{Bare}}
    & \multicolumn{2}{c}{\textbf{Explicit}} & \multicolumn{2}{c}{\textbf{Implicit}} & \multicolumn{2}{c}{\textbf{Bare}} \\
    \cmidrule(lr){2-3} \cmidrule(lr){4-5} \cmidrule(lr){6-7}
    \cmidrule(lr){8-9} \cmidrule(lr){10-11} \cmidrule(lr){12-13}
    \textbf{Method}
      & Recall & \textsc{CR} & Recall & \textsc{CR} & Recall & \textsc{CR}
      & Recall & \textsc{CR} & Recall & \textsc{CR} & Recall & \textsc{CR} \\
    \midrule
    StructGPT
      & 19.53 & 13.99  & 16.22 & 12.00  & 11.25 &  8.74
      & 38.44 & 25.81  & 29.97 & 20.14  & 18.82 & 12.99 \\
    ToG
      & 19.35 & 16.62  & 16.98 & 15.95  & 12.19 & 15.31
      & 21.59 & 19.42  & 18.95 & 19.47  & 12.19 & 16.71 \\
    ToG-2
      & 33.11 & 37.45  & 29.39 & \underline{38.30}  & 17.52 & \underline{32.57}
      & 42.03 & \underline{44.19}  & 32.21 & \underline{40.29}  & 18.55 & \underline{33.21} \\
    PoG
      & \underline{46.15} & \underline{38.55}  & \underline{38.04} & 32.91  & \underline{26.79} & 26.27
      & \underline{43.95} & 38.01  & \underline{38.40} & 33.59  & \underline{29.17} & 28.09 \\
    \rowcolor[HTML]{F4EBFF}
    \textbf{AdaPath}
      & \textbf{55.82} & \textbf{50.17}  & \textbf{48.79} & \textbf{48.39}  & \textbf{33.47} & \textbf{41.18}
      & \textbf{56.05} & \textbf{50.76}  & \textbf{48.34} & \textbf{47.63}  & \textbf{33.83} & \textbf{41.01} \\
    \bottomrule
  \end{tabularx}
  \caption{\textbf{Path-level Analysis on \sys{}.} We report Recall (\%) and Context Relevance (CR, \%) by query type
  and backbone, averaged across all hop depths.
  Per-column best \textbf{bolded}, second-best \underline{underlined}.}
  \label{tab:recall_ctx_relevance}
\end{table*}

\subsection{Depth-level Analysis of \sys{}}
\label{sec:analysis:depth}

\begin{figure*}[!t]
  \centering
  \includegraphics[width=0.95\textwidth]{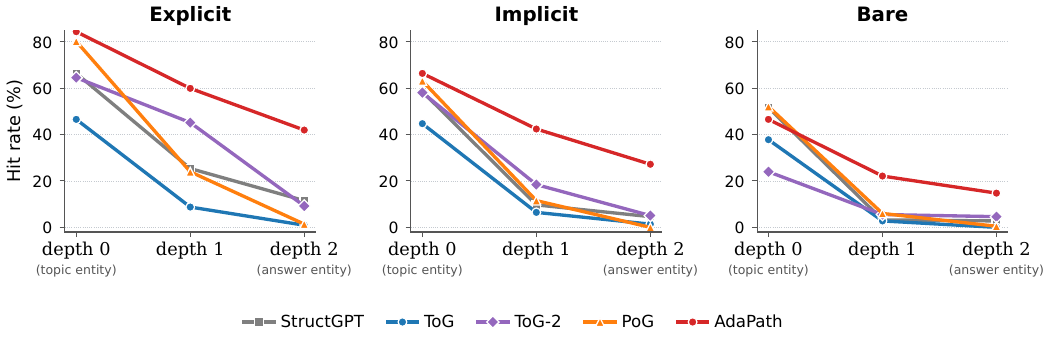}
  \caption{\textbf{Depth-level Analysis on \sys{}.} We report hit rate on 3-hop \textsc{BioStrat-QA} queries.
  Explicit (\textit{left}) / Implicit (\textit{middle}) / Bare (\textit{right}) panels plot the fraction of queries whose explored entity set contains the reference $d$-th node along the evidence path. Values shown are extracted  for Llama-3.1-70B. \sys{} sustains its hit rate at later depths while baselines drop sharply.}
  \label{fig:analysis_hit_ratio}
\end{figure*}

\begin{figure}[!t]
  \centering
  \includegraphics[width=\columnwidth]{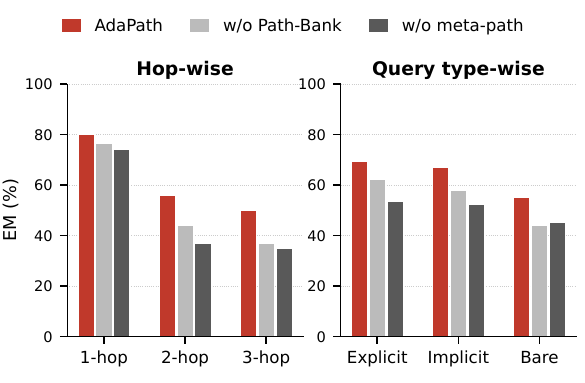}
  \caption{\textbf{Path-Bank ablation across hop depth and query type.}
  EM (\%) averaged across Llama-3.1-70B and Qwen-2.5-72B.
  Hop-wise (\textit{left}), averaged over query types.
  Query type-wise (\textit{right}), sample-size weighted average across hop depths.
  \textit{w/o Path-Bank} uses LLM-generated meta-path as \cradd{guidance} instead of using pre-defined Path-Bank; \textit{w/o meta-path} routes every query
  through free BFS with no Path-Bank guidance at all. 
  The results indicate the effectiveness of Path-Bank's query-adaptive meta-path during path-finding.}
  \label{fig:ablation_combined_v1}
\end{figure}

While the previous analysis examined path-level recovery, we also measure depth-level traversal accuracy on 3-hop queries by tracking whether each method reaches the correct entity at each reasoning depth in Figure~\ref{fig:analysis_hit_ratio}. 
\sys{} shows the highest hit rate at every depth across all query types in the dense biomedical KG, while other path-finding baselines, such as PoG and ToG-2, reach reasonable entities at the first hop but drop sharply at deeper depths. 
The robustness across depths explains \sys{}'s strong performance on multi-hop biomedical QA across all query types. 
On 2-hop queries, we observe a similar pattern, with \sys{} again sustaining the highest hit rate at later depths across all query types (Appendix~\ref{sec:appendix:perdepth2hop}).
We also conduct case studies to demonstrate the strength of \sys{} compared to baselines in multi-hop path-finding for \textit{implicit} and \textit{bare} biomedical queries (Appendix~\ref{sec:appendix:cases}).

\subsection{Ablation Study on \sys{}}
\label{sec:analysis:ablation}


To isolate the effect of query-adaptive meta-path guidance from Path-Bank, we compare \sys{} against two ablations on \textsc{BioStrat-QA} along two axes, hop depth and query type (Figure~\ref{fig:ablation_combined_v1}). 
The \textit{w/o Path-Bank} variant replaces the pre-defined Path-Bank with LLM-generated meta-paths, where the LLM plans node- and relation-level chains at inference time for each test query, isolating the effect of Path-Bank's offline mining. 
The \textit{w/o meta-path} variant removes meta-path guidance entirely. 
Both ablations fall below \sys{}, and the gap widens as queries span more hops. 
By query type, \sys{} outperforms both ablations across all types, and \textit{w/o Path-Bank} underperforms \textit{w/o meta-path} on bare queries where the LLM lacks sufficient surface cues to plan useful meta-paths, validating the quality of Path-Bank's query-adaptive meta-path selection.
We further provide \cradd{detailed} ablation results in \cradd{Appendix Tables~\ref{tab:appendix_ablation_biostrat} and~\ref{tab:appendix_ablation_starkmed}}.
\let\FloatBarrier\relax

%% file: sections/06_conclusion.tex
\section{Conclusion}
\label{sec:conclusion}



To address challenges in biomedical KGQA, we present \sys{}, a path-finding framework guided by query-adaptive meta-paths from a Path-Bank reflecting both query semantics and KG structure. 
We further release \textsc{BioStrat-QA}, a biomedical KGQA benchmark stratifying multi-hop queries by surface exposure of intermediate reasoning. 
\sys{} consistently outperforms baselines. Path- and depth-level analyses confirm that this advantage originates in the path-finding stage, where \sys{} retrieves paths quantitatively and qualitatively optimized for each query.

%% file: sections/07_limitations.tex

\sys{} relies on LLM scoring to select the next-hop frontier after query-adaptive meta-path filtering, so weaker backbones produce noisier per-hop decisions that propagate across hops and degrade longer reasoning chains.
\cradd{Meta-path filtering also depends on LLM-inferred hop length and answer type, and queries without a matching meta-path fall back to unguided traversal.}
The reachable answer space of \sys{} is bounded by what the underlying graph encodes, leaving queries that hinge on absent node or relation types, such as recently approved drugs or newly characterised mechanisms, difficult to answer through path-finding alone.
\cradd{Our benchmarks share a common synthetic construction procedure, and validation on independently collected biomedical QA remains future work. The mechanistically grounded paths \sys{} presents may also invite over-trust, and its answers should not substitute for expert judgment.}




%% file: sections/A_appendix.tex
\section{\textsc{BioStrat-QA} Dataset Details}
\label{sec:appendix:dataset}

This section details the construction of \textsc{BioStrat-QA}, covering
the split-wise statistics, the metapath template inventory, and the
verbatim prompts used to generate the \textsc{explicit},
\textsc{implicit}, and \textsc{bare} queries from a single evidence
path.

\subsection{Dataset Statistics}
\label{sec:appendix:dataset:stats}

\textsc{BioStrat-QA} contains $4{,}568$ queries split into train, dev,
and test with a roughly $5{:}2{:}3$ ratio, jointly covering 1-, 2-, and
3-hop biomedical reasoning over PrimeKG
(Table~\ref{tab:appendix_dataset_stats}). Each record carries the same
topic and answer entities under three query formulations, so per-record
difficulty is controlled by information level rather than by a change of
topic or answer.

\begin{table}[!htbp]
  \centering\small
  \setlength{\tabcolsep}{6pt}
  \renewcommand{\arraystretch}{1.1}
  \begin{tabular*}{\columnwidth}{@{\extracolsep{\fill}}lrrrr@{}}
    \toprule
    \textbf{Split} & \textbf{Total} & \textbf{1-hop} & \textbf{2-hop} & \textbf{3-hop} \\
    \midrule
    train  & 2{,}491 &   922 & 1{,}119 & 450 \\
    dev    &   898   &   331 &    404  & 163 \\
    test   & 1{,}179 &   437 &    525  & 217 \\
    \midrule
    Total  & \textbf{4{,}568} & \textbf{1{,}690} & \textbf{2{,}048} & \textbf{830} \\
    \bottomrule
  \end{tabular*}
  \caption{\textbf{\textsc{BioStrat-QA} size by split and hop depth.}
  Each record carries three query formulations, explicit, implicit, and
  bare, over the same evidence path.}
  \label{tab:appendix_dataset_stats}
\end{table}

\subsection{Metapath Templates}
\label{sec:appendix:dataset:templates}

The template inventory takes the STaRK-Prime~\citep{wu2024stark}
templates as its core and extends them so that \textsc{BioStrat-QA}
covers reasoning paths up to three hops. Queries are generated from $60$
hand-crafted metapath templates over PrimeKG, consisting of $18$ one-hop,
$30$ two-hop, and $12$ three-hop patterns. Most templates are
single-topic chains. A subset encodes a junction pattern where two named
entities meet at a shared intermediate or answer node.
Tables~\ref{tab:appendix_templates_1hop},
\ref{tab:appendix_templates_2hop}, and~\ref{tab:appendix_templates_3hop}
list every template.

\begin{table}[!htbp]
  \centering\footnotesize
  \setlength{\tabcolsep}{4pt}
  \renewcommand{\arraystretch}{1.1}
  \begin{tabularx}{\columnwidth}{@{}l >{\raggedright\arraybackslash}X@{}}
    \toprule
    \textbf{ID} & \textbf{Metapath} \\
    \midrule
    1h-01 & drug $\to$[indication]$\to$ disease \\
    1h-02 & drug $\to$[side effect]$\to$ effect/phenotype \\
    1h-03 & drug $\to$[target]$\to$ gene/protein \\
    1h-04 & disease $\to$[associated with]$\to$ gene/protein \\
    1h-05 & disease $\to$[phenotype present]$\to$ effect/phenotype \\
    1h-06 & gene/protein $\to$[expression present]$\to$ anatomy \\
    1h-07 & drug $\to$[contraindication]$\to$ disease \\
    1h-08 & gene/protein $\to$[ppi]$\to$ gene/protein \\
    1h-09 & drug $\to$[enzyme]$\to$ gene/protein \\
    1h-10 & effect/phenotype $\to$[associated with]$\to$ gene/protein \\
    1h-11 & effect/phenotype $\to$[phenotype absent]$\to$ disease \\
    1h-12 & drug $\to$[contraindication]$\to$ disease $\leftarrow$[associated with]$\leftarrow$ gene/protein\textsuperscript{$\dagger$} \\
    1h-13 & anatomy $\to$[expression present]$\to$ gene/protein $\leftarrow$[expression absent]$\leftarrow$ anatomy\textsuperscript{$\dagger$} \\
    1h-15 & drug $\to$[carrier]$\to$ gene/protein $\leftarrow$[carrier]$\leftarrow$ drug\textsuperscript{$\dagger$} \\
    1h-16 & drug $\to$[off-label use]$\to$ disease \\
    1h-17 & exposure $\to$[linked to]$\to$ disease \\
    1h-18 & drug $\to$[transporter]$\to$ gene/protein \\
    1h-19 & gene/protein $\to$[interacts with]$\to$ biological\_process \\
    \bottomrule
  \end{tabularx}
  \caption{\textbf{1-hop metapath templates.}
  $\dagger$~marks a 2-topic junction template.}
  \label{tab:appendix_templates_1hop}
\end{table}

\begin{table*}[!htbp]
  \centering\footnotesize
  \setlength{\tabcolsep}{4pt}
  \renewcommand{\arraystretch}{1.1}
  \begin{tabularx}{\textwidth}{@{}l >{\raggedright\arraybackslash}X@{}}
    \toprule
    \textbf{ID} & \textbf{Metapath} \\
    \midrule
    2h-01 & anatomy $\to$[expression present]$\to$ gene/protein $\to$[target]$\to$ drug \\
    2h-02 & drug $\to$[side effect]$\to$ effect/phenotype $\to$[side effect]$\to$ drug \\
    2h-03 & drug $\to$[carrier]$\to$ gene/protein $\to$[carrier]$\to$ drug \\
    2h-04 & anatomy $\to$[expression present]$\to$ gene/protein $\to$[enzyme]$\to$ drug \\
    2h-05 & cellular\_component $\to$[interacts with]$\to$ gene/protein $\to$[carrier]$\to$ drug \\
    2h-06 & molecular\_function $\to$[interacts with]$\to$ gene/protein $\to$[target]$\to$ drug \\
    2h-07 & effect/phenotype $\to$[side effect]$\to$ drug $\to$[synergistic interaction]$\to$ drug \\
    2h-08 & disease $\to$[indication]$\to$ drug $\to$[contraindication]$\to$ disease \\
    2h-09 & disease $\to$[parent-child]$\to$ disease $\to$[phenotype present]$\to$ effect/phenotype \\
    2h-10 & gene/protein $\to$[transporter]$\to$ drug $\to$[side effect]$\to$ effect/phenotype \\
    2h-11 & drug $\to$[transporter]$\to$ gene/protein $\to$[interacts with]$\to$ exposure \\
    2h-12 & pathway $\to$[interacts with]$\to$ gene/protein $\to$[ppi]$\to$ gene/protein \\
    2h-13 & drug $\to$[synergistic interaction]$\to$ drug $\to$[transporter]$\to$ gene/protein \\
    2h-15 & effect/phenotype $\to$[associated with]$\to$ gene/protein $\to$[interacts with]$\to$ biological\_process \\
    2h-16 & drug $\to$[transporter]$\to$ gene/protein $\to$[expression present]$\to$ anatomy \\
    2h-17 & drug $\to$[target]$\to$ gene/protein $\to$[interacts with]$\to$ cellular\_component \\
    2h-18 & biological\_process $\to$[interacts with]$\to$ gene/protein $\to$[expression absent]$\to$ anatomy \\
    2h-19 & effect/phenotype $\to$[associated with]$\to$ gene/protein $\to$[expression absent]$\to$ anatomy \\
    2h-20 & drug $\to$[indication]$\to$ disease $\to$[indication]$\to$ drug\textsuperscript{$\ast$} \\
    2h-22 & gene/protein $\to$[associated with]$\to$ disease $\to$[associated with]$\to$ gene/protein\textsuperscript{$\ast$} \\
    2h-23 & gene/protein $\to$[associated with]$\to$ effect/phenotype $\to$[associated with]$\to$ gene/protein\textsuperscript{$\ast$} \\
    2h-25 & disease $\to$[associated with]$\to$ gene/protein $\to$[carrier]$\to$ drug $\leftarrow$[side effect]$\leftarrow$ effect/phenotype\textsuperscript{$\dagger$} \\
    2h-26 & drug $\to$[target]$\to$ gene/protein $\to$[associated with]$\to$ disease $\leftarrow$[phenotype present]$\leftarrow$ effect/phenotype\textsuperscript{$\dagger$} \\
    2h-28 & drug $\to$[off-label use]$\to$ disease $\to$[associated with]$\to$ gene/protein \\
    2h-29 & exposure $\to$[linked to]$\to$ disease $\to$[indication]$\to$ drug \\
    2h-30 & drug $\to$[indication]$\to$ disease $\to$[phenotype absent]$\to$ effect/phenotype \\
    2h-31 & disease $\to$[phenotype absent]$\to$ effect/phenotype $\to$[associated with]$\to$ gene/protein \\
    2h-33 & drug $\to$[synergistic interaction]$\to$ drug $\to$[target]$\to$ gene/protein \\
    2h-34 & disease $\to$[parent-child]$\to$ disease $\to$[associated with]$\to$ gene/protein \\
    2h-35 & drug $\to$[transporter]$\to$ gene/protein $\to$[associated with]$\to$ disease \\
    \bottomrule
  \end{tabularx}
  \caption{\textbf{2-hop metapath templates.}
  $\ast$~marks a conjoint template with an auxiliary direct constraint
  between the two ends, such as a synergistic interaction or a PPI.
  $\dagger$~marks a 2-topic junction template.}
  \label{tab:appendix_templates_2hop}
\end{table*}

\begin{table*}[!htbp]
  \centering\footnotesize
  \setlength{\tabcolsep}{4pt}
  \renewcommand{\arraystretch}{1.1}
  \begin{tabularx}{\textwidth}{@{}l >{\raggedright\arraybackslash}X@{}}
    \toprule
    \textbf{ID} & \textbf{Metapath} \\
    \midrule
    3h-03 & disease $\to$[associated with]$\to$ gene/protein $\to$[target]$\to$ drug $\to$[side effect]$\to$ effect/phenotype \\
    3h-04 & disease $\to$[associated with]$\to$ gene/protein $\to$[ppi]$\to$ gene/protein $\to$[target]$\to$ drug \\
    3h-05 & drug $\to$[target]$\to$ gene/protein $\to$[associated with]$\to$ disease $\to$[phenotype present]$\to$ effect/phenotype \\
    3h-07 & effect/phenotype $\to$[associated with]$\to$ gene/protein $\to$[target]$\to$ drug $\to$[indication]$\to$ disease \\
    3h-08 & gene/protein $\to$[associated with]$\to$ disease $\to$[indication]$\to$ drug $\to$[side effect]$\to$ effect/phenotype \\
    3h-09 & drug $\to$[carrier]$\to$ gene/protein $\to$[associated with]$\to$ disease $\to$[phenotype present]$\to$ effect/phenotype \\
    3h-10 & disease $\to$[indication]$\to$ drug $\to$[target]$\to$ gene/protein $\to$[expression present]$\to$ anatomy \\
    3h-12 & disease $\to$[associated with]$\to$ gene/protein $\to$[target]$\to$ drug $\to$[side effect]$\to$ effect/phenotype $\leftarrow$[associated with]$\leftarrow$ gene/protein\textsuperscript{$\dagger$} \\
    3h-18 & exposure $\to$[linked to]$\to$ disease $\to$[indication]$\to$ drug $\to$[side effect]$\to$ effect/phenotype \\
    3h-19 & drug $\to$[indication]$\to$ disease $\to$[phenotype absent]$\to$ effect/phenotype $\to$[associated with]$\to$ gene/protein \\
    3h-22 & drug $\to$[target]$\to$ gene/protein $\to$[transporter]$\to$ drug $\to$[indication]$\to$ disease \\
    3h-23 & biological\_process $\to$[interacts with]$\to$ gene/protein $\to$[ppi]$\to$ gene/protein $\to$[expression present]$\to$ anatomy \\
    \bottomrule
  \end{tabularx}
  \caption{\textbf{3-hop metapath templates.}
  $\dagger$~marks a 2-topic junction template.}
  \label{tab:appendix_templates_3hop}
\end{table*}

\subsection{Query Generation Prompts}
\label{sec:appendix:dataset:prompts}

Given an evidence path, three queries are generated through separate
GPT-5.4 batch calls, one per information level, so that no information
leaks across formulations of the same path. The \textsc{implicit} level
additionally takes a precomputed pathway summary that paraphrases
intermediate node and relation descriptions without naming any
intermediate or answer entity. The verbatim prompts are listed below.

\paragraph{Pathway summary (used as input to \textsc{implicit} for
2- and 3-hop paths).}

\begin{tcolorbox}[colback=gray!4,colframe=gray!40,boxrule=0.5pt,
                  arc=2pt,left=4pt,right=4pt,top=3pt,bottom=3pt,
                  fontupper=\footnotesize\ttfamily,breakable]
\textbf{System prompt}\\[2pt]
You are a biomedical knowledge expert. You are given a biomedical
pathway from a knowledge graph with node names, descriptions, and
relations.\\[4pt]
Your task: Summarize the biomedical pathway in 2-4 sentences.\\[4pt]
Rules:\\
- In your summary, do NOT directly use the names of intermediate nodes
or the answer node. Paraphrase them based on the provided descriptions.\\
- The summary should read as a natural biomedical explanation of how
the topic entity relates to an entity of the answer type.
\end{tcolorbox}

The user message provides the topic entity with its name and type, each
hop in the form \textit{source name}, \textit{relation}, \textit{target
description}, and the type of the target entity without its name. For
1-hop paths this step is skipped and a hand-written relation-level
description is used directly.

\paragraph{Explicit query.}
The model sees the full path with intermediate names and relation
labels, and is told not to reveal the answer entity name.

\begin{tcolorbox}[colback=gray!4,colframe=gray!40,boxrule=0.5pt,
                  arc=2pt,left=4pt,right=4pt,top=3pt,bottom=3pt,
                  fontupper=\footnotesize\ttfamily,breakable]
\textbf{System prompt}\\[2pt]
You are a biomedical expert. Generate a natural biomedical question
based on the provided evidence path.\\[4pt]
Rules:\\
- Use the topic entity name.\\
- Use intermediate node names from the evidence path.\\
- Use relation names from the evidence path (e.g., ``indication'', ``target'').\\
- Use the answer description (if provided) to specify the answer, but
do NOT mention the answer entity name. If no answer description is
provided, you may use your biomedical knowledge about the answer
entity to add specificity.\\
- The question should sound natural, as if asked by a researcher or
clinician.\\
- Output ONLY the question, nothing else.\\[6pt]
\textbf{User prompt template}\\[2pt]
Topic entity: \textless topic\_name\textgreater\ (\textless topic\_type\textgreater)\\[4pt]
Evidence path:\\
\textless hop1.src\textgreater\ -[\textless hop1.rel\textgreater]-\textgreater\ \textless hop1.dst\textgreater\\
\textless hop2.src\textgreater\ -[\textless hop2.rel\textgreater]-\textgreater\ \textless hop2.dst\textgreater\\
...\\[4pt]
Answer entity: \textless answer\_name\textgreater\ (\textless answer\_type\textgreater)\\
Answer description: \textless answer\_desc\textgreater\\[4pt]
Generate a question that follows this evidence path.\\
Do NOT mention the answer entity name ``\textless answer\_name\textgreater''
in the question.
\end{tcolorbox}

\paragraph{Implicit query.}
Intermediate names and relation labels are withheld. The model instead
receives the pathway summary for 2- and 3-hop paths, or a canonical
relation description for 1-hop paths.

\begin{tcolorbox}[colback=gray!4,colframe=gray!40,boxrule=0.5pt,
                  arc=2pt,left=4pt,right=4pt,top=3pt,bottom=3pt,
                  fontupper=\footnotesize\ttfamily,breakable]
\textbf{System prompt}\\[2pt]
You are a biomedical expert. Generate a natural biomedical question
based on the provided pathway information.\\[4pt]
Rules:\\
- Use the topic entity name.\\
- Do NOT mention intermediate node names directly.\\
- Do NOT use relation names directly. Paraphrase based on the relation
context (1-hop) or pathway summary (2-3 hop).\\
- Use the answer description (if provided) to specify the answer, but
do NOT mention the answer entity name. If no answer description is
provided, you may use your biomedical knowledge about the answer
entity to add specificity.\\
- The question should sound natural, as if asked by a researcher or
clinician.\\
- Output ONLY the question, nothing else.\\[6pt]
\textbf{User prompt template}\\[2pt]
Topic entity: \textless topic\_name\textgreater\ (\textless topic\_type\textgreater)\\[4pt]
\# 1-hop: relation description from relation\_descriptions.json\\
Relation context: \textless relation\_desc\textgreater\\[4pt]
\# OR 2-3 hop: pathway summary from the previous stage\\
Pathway summary:\\
\textless name-free 2-4 sentence summary\textgreater\\[4pt]
Answer entity: \textless answer\_name\textgreater\ (\textless answer\_type\textgreater)\\
Answer description: \textless answer\_desc\textgreater\\[4pt]
Generate a question reflecting the pathway above.\\
Do NOT mention intermediate node names, relation names, or the answer
entity name ``\textless answer\_name\textgreater'' in the question.
\end{tcolorbox}

\paragraph{Bare query.}
No path detail is shown, and the model is constrained to use a generic
linking phrase rather than a specific relation verb.

\begin{tcolorbox}[colback=gray!4,colframe=gray!40,boxrule=0.5pt,
                  arc=2pt,left=4pt,right=4pt,top=3pt,bottom=3pt,
                  fontupper=\footnotesize\ttfamily,breakable]
\textbf{System prompt}\\[2pt]
You are a biomedical expert. Generate a natural biomedical question
that asks about a relationship between the given entities.\\[4pt]
Rules:\\
- Use the topic entity name.\\
- Do NOT mention intermediate node names or any pathway/mechanism details.\\
- Do NOT use specific relation verbs (e.g., ``treats'', ``targets'',
``induces'', ``expressed in''). Use generic linking phrases like
``associated with'', ``related to'', ``linked to'', ``connected to'',
or natural variations appropriate for the entity types.\\
- Use the answer description (if provided) to specify the answer, but
do NOT mention the answer entity name. If no answer description is
provided, you may use your biomedical knowledge about the answer
entity to add specificity.\\
- Use varied and natural phrasing.\\
- Output ONLY the question, nothing else.\\[6pt]
\textbf{User prompt template}\\[2pt]
Topic entity: \textless topic\_name\textgreater\ (\textless topic\_type\textgreater)\\[4pt]
Answer entity: \textless answer\_name\textgreater\ (\textless answer\_type\textgreater)\\
Answer description: \textless answer\_desc\textgreater\\[4pt]
Generate a natural question about a \textless answer\_type\textgreater\
that has some relationship with \textless topic\_name\textgreater,
using varied and natural phrasing.\\
Do NOT mention the answer entity name
``\textless answer\_name\textgreater'' or any specific
pathway/mechanism details.
\end{tcolorbox}

\section{Results on Small Backbones}
\label{sec:appendix:extended}

We report \sys{} and baseline results on smaller backbones, Llama-3.1-8B
and Qwen-2.5-7B, complementing the large-backbone main results in
Section~\ref{sec:experiments:main}
(Tables~\ref{tab:appendix_biotog_small}
and~\ref{tab:appendix_stark_medddx_small}).
Performance levels shift with backbone scale, yet \sys{} retains the
pattern of improvement over baselines observed with the larger backbones.
\begin{table*}[!htbp]
  \centering\footnotesize
  \setlength{\tabcolsep}{4pt}
  \setlength{\dashlinedash}{0.5pt}
  \setlength{\dashlinegap}{1.5pt}
  \renewcommand{\arraystretch}{1.05}
  \resizebox{\textwidth}{!}{
  \begin{tabular}{@{}ll cccc cccc cccc c@{}}
    \toprule
    \multirow{2}{*}{\textbf{Category}} & \multirow{2}{*}{\textbf{Method}} & \multicolumn{4}{c}{\textbf{Explicit}} & \multicolumn{4}{c}{\textbf{Implicit}} & \multicolumn{4}{c}{\textbf{Bare}} & \multirow{2}{*}{\textbf{Overall}} \\
    \cmidrule(lr){3-6} \cmidrule(lr){7-10} \cmidrule(lr){11-14}
    & & 1-hop & 2-hop & 3-hop & Avg & 1-hop & 2-hop & 3-hop & Avg & 1-hop & 2-hop & 3-hop & Avg & \\
    \midrule
    \rowcolor{lightgray}
    \multicolumn{15}{c}{\texttt{Llama-3.1-8B}} \\
    \midrule
    \multirow{3}{*}{\textbf{LLM only}}
    & IO        & 32.7 & 24.6 & 19.4 & 26.6 & 31.6 & 27.8 & 19.8 & 27.7 & 33.0 & 25.1 & 22.1 & 27.5 & 27.3 \\
    & CoT       & 37.1 & 29.7 & 21.2 & 30.9 & 32.7 & 29.5 & 26.7 & 30.2 & 33.2 & 25.9 & \underline{24.9} & 28.4 & 29.8 \\
    & SC        & 34.8 & 28.2 & 21.7 & 29.4 & 30.9 & 30.3 & \underline{27.6} & 30.0 & 33.2 & 25.5 & 24.9 & 28.2 & 29.2 \\

    \noalign{\vskip 2pt}    \cdashline{1-15}    \noalign{\vskip 2pt}

    \multirow{2}{*}{\textbf{Fact-retrieval}}
    & CoK        & 37.3 & \underline{30.5} & \underline{23.0} & 31.6 & 35.7 & \underline{31.8} & 27.2 & 32.4 & 38.2 & \underline{28.6} & \textbf{25.4} & 31.6 & 31.9 \\
    & KGARevion  & 30.9 & 26.9 & 20.7 & 27.2 & 31.1 & 31.2 & \underline{27.6} & 30.5 & 31.1 & 21.7 & 23.0 & 25.4 & 27.7 \\

    \noalign{\vskip 2pt}    \cdashline{1-15}    \noalign{\vskip 2pt}

    \multirow{5}{*}{\textbf{Path-finding}}
    & StructGPT  & 43.7 & 22.7 & 17.5 & 29.5 & 36.2 & 25.9 & 18.4 & 28.3 & 33.2 & 17.3 & 15.2 & 22.8 & 26.9 \\
    & ToG        & 46.9 & 28.0 & 18.9 & 33.3 & 41.2 & 25.3 & 19.8 & 30.2 & 36.4 & 15.8 & 14.7 & 23.2 & 28.9 \\
    & ToG-2      & 64.1 & 22.1 & 18.0 & \underline{36.9} & \underline{58.1} & 23.6 & 18.4 & \underline{35.5} & \textbf{57.9} & 20.6 & 15.7 & \underline{33.5} & \underline{35.3} \\
    & PoG        & \underline{67.7} & 9.0 & 7.8 & 30.5 & 53.3 & 5.7 & 7.8 & 23.7 & 35.7 & 2.7 & 9.2 & 16.1 & 23.5 \\
    \rowcolor[HTML]{F4EBFF}
    \cellcolor{white} & \textbf{\sys{}}
                  & \textbf{74.4} & \textbf{47.8} & \textbf{28.6} & \textbf{54.1} & \textbf{76.2} & \textbf{44.6} & \textbf{32.3} & \textbf{54.0} & \underline{56.3} & \textbf{31.6} & \textbf{25.4} & \textbf{39.6} & \textbf{49.2} \\
    \midrule
    \rowcolor{lightgray}
    \multicolumn{15}{c}{\texttt{Qwen-2.5-7B}} \\
    \midrule
    \multirow{3}{*}{\textbf{LLM only}}
    & IO        & 33.4 & 21.9 & 19.8 & 25.8 & 27.7 & 25.3 & 21.2 & 25.4 & 26.8 & 19.2 & 19.8 & 22.1 & 24.4 \\
    & CoT       & 29.7 & 22.9 & 21.2 & 25.1 & 25.4 & 23.8 & 30.9 & 25.7 & 25.4 & 23.4 & 24.9 & 24.4 & 25.1 \\
    & SC        & 31.8 & 22.7 & 21.7 & 25.9 & 26.3 & \underline{27.8} & \underline{31.3} & 27.9 & 26.1 & \underline{23.8} & 29.0 & 25.6 & 26.5 \\

    \noalign{\vskip 2pt}    \cdashline{1-15}    \noalign{\vskip 2pt}

    \multirow{2}{*}{\textbf{Fact-retrieval}}
    & CoK        & 33.4 & 23.2 & 21.7 & 26.7 & 29.1 & 27.0 & 30.4 & 28.4 & 31.4 & 21.7 & \textbf{30.4} & 26.9 & 27.3 \\
    & KGARevion  & 25.9 & 18.7 & 22.1 & 22.0 & 23.3 & 23.2 & 21.2 & 22.9 & 21.5 & 17.5 & 18.0 & 19.1 & 21.3 \\

    \noalign{\vskip 2pt}    \cdashline{1-15}    \noalign{\vskip 2pt}

    \multirow{5}{*}{\textbf{Path-finding}}
    & StructGPT  & 60.6 & 21.1 & 19.8 & 35.5 & 53.8 & 19.6 & 23.5 & 33.0 & 38.0 & 10.1 & 11.5 & 20.7 & 29.7 \\
    & ToG        & 47.4 & 24.0 & 17.1 & 31.4 & 44.4 & 23.4 & 24.0 & 31.3 & 38.0 & 14.9 & 16.1 & 23.7 & 28.8 \\
    & ToG-2      & \textbf{73.0} & \underline{29.1} & \underline{29.0} & \underline{45.4} & \underline{63.4} & 25.9 & 19.4 & \underline{38.6} & \underline{56.5} & 18.5 & 14.8 & \underline{31.9} & \underline{38.6} \\
    & PoG        & 63.6 & 14.9 & 14.3 & 32.8 & 60.2 & 20.2 & 21.2 & 35.2 & 45.8 & 9.1 & 14.3 & 23.7 & 30.6 \\
    \rowcolor[HTML]{F4EBFF}
    \cellcolor{white} & \textbf{\sys{}}
                  & \underline{72.5} & \textbf{52.2} & \textbf{43.8} & \textbf{58.2} & \textbf{70.2} & \textbf{51.0} & \textbf{34.6} & \textbf{55.1} & \textbf{60.6} & \textbf{37.7} & \underline{29.9} & \textbf{44.8} & \textbf{52.7} \\
    \bottomrule
  \end{tabular}
  }
  \caption{\textbf{Main results on \textsc{BioStrat-QA}, small backbones.}
  EM (\%) per (information-level, hop-depth) cell; \textbf{Overall}
  averages all nine cells.
  Per-column best is \textbf{bold} and second-best is \underline{underlined}
  within each backbone block.}
  \label{tab:appendix_biotog_small}
\end{table*}

\section{Topic Entity Extraction for STaRK-Prime and MedDDx}
\label{sec:appendix:topic_extraction}

STaRK-Prime~\citep{wu2024stark} and MedDDx share the same underlying
query set, so we run topic entity extraction once and reuse the
resulting entity links across both evaluations. The extraction follows a
four-stage pipeline that interleaves typed graph traversal with two
GPT-5.4 batch calls, narrowing the candidate metapaths and all
type-matching nodes in PrimeKG down to one template and one entity per
topic slot.

\paragraph{Stage 1: Typed reverse BFS feasibility filter.}
The search space is a set of metapath templates built on the STaRK-Prime
template set~\citep{wu2024stark}, complemented with additional
single-edge patterns enumerated from the PrimeKG schema. For each query
we identify which of these templates can actually reach the given answer
entity in PrimeKG. From every answer node we walk the template metapath
backward, enforcing edge type and target node type at every hop. For
queries with multiple valid answers, we intersect the candidate sets
obtained from each answer node, and fall back to their union when the
intersection is empty. Templates with two topic slots are traversed once
per slot. A template survives only if its reverse traversal returns at
least one candidate. A median of $\sim$7 templates survive per query and
are passed to the next stage.

\paragraph{Stage 2: LLM template pick.}
The surviving templates are passed to GPT-5.4 via the OpenAI batch API.
We prompt the model with the query, the answer entity, and the list of
graph-feasible templates, each shown as both its formal metapath and a
short paraphrase, and ask it to return the single template that best
matches the question's intent in strict JSON. Templates that fail to
parse or fall outside the allowed range are discarded.

\begin{table}[!htbp]
  \centering\footnotesize
  \setlength{\tabcolsep}{3pt}
  \setlength{\dashlinedash}{0.5pt}
  \setlength{\dashlinegap}{1.5pt}
  \renewcommand{\arraystretch}{1.05}
  \resizebox{\columnwidth}{!}{
  \begin{tabular}{@{}l cc cccc@{}}
    \toprule
    \multirow{2}{*}{\textbf{Method}} & \multicolumn{2}{c}{\textbf{STaRK-Prime}} & \multicolumn{4}{c}{\textbf{MedDDx}} \\
    \cmidrule(lr){2-3} \cmidrule(lr){4-7}
    & Synthesized & \begin{tabular}{@{}c@{}}Human-\\generated\end{tabular} & Basic & Inter. & Expert & All \\
    \midrule
    \rowcolor{lightgray}
    \multicolumn{7}{c}{\texttt{Llama-3.1-8B}} \\
    \midrule
    IO         & 13.1 & 12.2 & 42.9 & 36.2 & 34.6 & 36.7 \\
    CoT        & 13.1 & 15.3 & 41.6 & 32.9 & 37.9 & 35.5 \\
    SC         & 12.2 & 12.2 & 44.9 & 35.7 & 38.1 & 37.6 \\

    \noalign{\vskip 2pt}    \cdashline{1-7}    \noalign{\vskip 2pt}

    CoK        & 13.1 & 14.3 & 48.2 & 38.1 & 36.9 & 39.2 \\
    KGARevion  & 12.5 & 12.2 & 38.4 & 34.5 & 31.5 & 34.2 \\

    \noalign{\vskip 2pt}    \cdashline{1-7}    \noalign{\vskip 2pt}

    StructGPT  & 17.2 & 18.4 & 44.5 & 37.4 & 37.1 & 38.3 \\
    ToG        & 28.9 & 24.5 & \underline{51.8} & \underline{41.4} & \underline{40.8} & \underline{42.7} \\
    ToG-2      & \underline{29.2} & \underline{35.7} & 41.6 & 39.2 & 34.0 & 38.1 \\
    PoG        & 23.1 & 20.4 & 32.2 & 30.8 & 30.6 & 31.0 \\
    \rowcolor[HTML]{F4EBFF} \textbf{\sys{}}
               & \textbf{43.0} & \textbf{44.9} & \textbf{58.4} & \textbf{49.4} & \textbf{49.5} & \textbf{50.6} \\
    \midrule
    \rowcolor{lightgray}
    \multicolumn{7}{c}{\texttt{Qwen-2.5-7B}} \\
    \midrule
    IO         & 10.9 & 12.2 & 37.1 & 33.4 & 32.9 & 33.8 \\
    CoT        & 10.1 & 9.2 & 34.3 & 32.1 & 33.5 & 32.8 \\
    SC         & 10.5 & 9.2 & 40.0 & 33.1 & 30.8 & 33.5 \\

    \noalign{\vskip 2pt}    \cdashline{1-7}    \noalign{\vskip 2pt}

    CoK        & 11.0 & 9.2 & 36.3 & 32.3 & 34.8 & 33.5 \\
    KGARevion  & 10.6 & 4.1 & 37.1 & 32.1 & 32.1 & 32.8 \\

    \noalign{\vskip 2pt}    \cdashline{1-7}    \noalign{\vskip 2pt}

    StructGPT  & 25.3 & 23.5 & \underline{48.2} & 37.7 & 39.1 & 39.5 \\
    ToG        & 29.6 & 24.5 & 42.9 & 40.5 & 37.5 & 40.0 \\
    ToG-2      & \underline{31.2} & \underline{33.7} & 36.3 & 31.5 & 27.9 & 31.2 \\
    PoG        & 29.7 & 27.6 & 44.1 & \underline{41.5} & \underline{44.1} & \underline{42.6} \\
    \rowcolor[HTML]{F4EBFF} \textbf{\sys{}}
               & \textbf{43.7} & \textbf{37.8} & \textbf{50.6} & \textbf{47.6} & \textbf{44.5} & \textbf{47.2} \\
    \bottomrule
  \end{tabular}
  }
  \caption{\textbf{Results on STaRK-Prime and MedDDx, small backbones.}
  STaRK-Prime: EM (\%). MedDDx: accuracy (\%). Per-column best is \textbf{bold},
  second-best is \underline{underlined} within each backbone block.}
  \label{tab:appendix_stark_medddx_small}
\end{table}

\paragraph{Stage 3: Per-slot candidate scoring.}
For each topic slot of the chosen template we collect all PrimeKG nodes
whose type matches the slot, then rank them against the query by
combining BM25~\citep{lin2022pretrained} over name-plus-description text
with a sentence-BERT~\citep{reimers2019sentence} similarity. The two
scores are min--max normalised within the candidate pool and summed, and
the top-$10$ candidates are kept. To bound compute for slots with very
large candidate pools such as all gene/protein nodes, we first take the
top-$1{,}000$ by BM25 and then rerank with SBERT, which preserves
top-$10$ quality at a fraction of the cost. Each surviving candidate is
also annotated with the line-level chunk of its description that
maximises BM25 against the query, truncated to $300$ characters, which
provides compact evidence for the next stage.

\paragraph{Stage 4: Per-slot LLM entity pick.}
GPT-5.4 then selects the final topic entity from the top-$10$ candidates
per slot. The prompt presents the query, the answer entity, the slot's
type, and each candidate with its name and best chunk, and asks the
model to pick the entity the query starts from rather than the one it
seeks as the answer. Each candidate carries a node identifier so that
the model can disambiguate same-name nodes.

For each query the pipeline records the topic entity of every slot in the
selected template. The topic entity is passed unchanged to \sys{} and to
every path-finding baseline as the starting point for traversal on
STaRK-Prime and MedDDx.

\section{Evaluation Metrics}
\label{sec:appendix:metrics}

We summarise the four metrics used throughout the paper, namely exact
match (EM) on \textsc{BioStrat-QA} and STaRK-Prime, accuracy on MedDDx,
triplet-level recall on retrieved reasoning chains, and an LLM-judged
context-relevance score for retrieved paths.

\begin{figure*}[!htbp]
  \centering
  \includegraphics[width=0.95\textwidth]{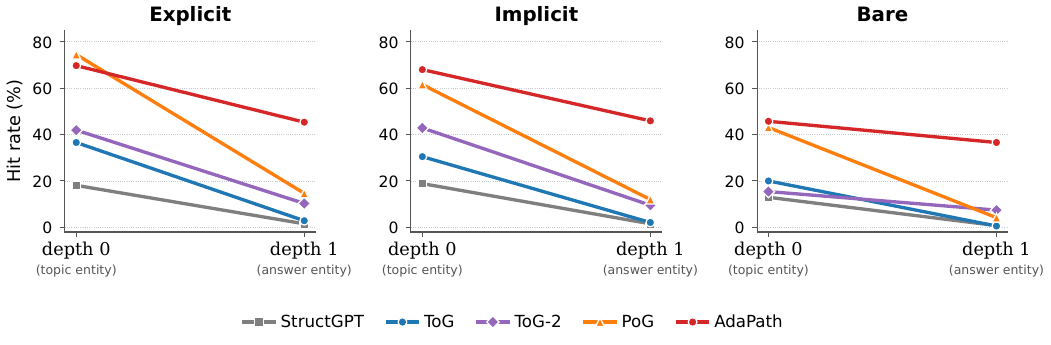}
  \caption{\textbf{Per-depth hit rate on 2-hop \textsc{BioStrat-QA} queries
  (Llama-3.1-70B).}
  Explicit (\textit{left}), Implicit (\textit{middle}), and Bare
  (\textit{right}) panels plot the fraction of queries whose explored
  entity set contains the reference $d$-th node along the evidence path.
  Each subplot uses an independent $y$-axis. \sys{} sustains its hit rate
  at the answer depth across all three query formulations, while
  path-finding baselines drop sharply.}
  \label{fig:appendix_perdepth_2hop}
\end{figure*}

\paragraph{Exact Match (EM).}
For \textsc{BioStrat-QA} and STaRK-Prime, a prediction is counted correct
when the model's final answer matches any of the ground-truth answer
names recorded for the query, with case and whitespace normalised. Each
record is released with a multi-answer expansion, so the model is
credited for identifying any of the semantically equivalent answer
entities rather than having to recover a single canonical name. EM is
reported as a percentage and, for \textsc{BioStrat-QA}, micro-averaged
within each cell.

\paragraph{Accuracy.}
For the multiple-choice queries in MedDDx, we report standard
classification accuracy. A prediction is counted correct when it matches
either the ground-truth option letter or the name of the entity that
option refers to.

\paragraph{Recall.}
For path-finding methods we evaluate path recovery via per-depth triplet
recall against the ground-truth (GT) evidence path. Let
$g_d = (h_d, r_d, t_d)$ denote the GT triplet at reasoning depth $d$ and
$P_d$ the set of retrieved triplets at depth $d$, capped at $K = 100$
random samples per depth with seed $42$. We compare triplets after
lowercasing the head and tail entity names, writing
$\mathrm{norm}(h, r, t) = (\mathrm{lower}(h), r, \mathrm{lower}(t))$, and
the hit indicator at depth $d$ is
\begin{equation}
  \mathrm{Hit}_d = \mathbf{1}\!\left[\mathrm{norm}(g_d) \in
      \{\mathrm{norm}(p) : p \in P_d\}\right].
\end{equation}
Per-record recall is the depth-wise average of $\mathrm{Hit}_d$ over the
$D$ depths of that record. The reported recall is the micro-average over
all record and depth pairs across the 1-, 2-, and 3-hop subsets, dividing
the total number of hits by the total number of pairs.

\paragraph{Context Relevance.}
To complement triplet-exact recall, which penalises retrieved paths that
pass through semantically valid neighbours without exactly reproducing
the GT triplets, we additionally report Context Relevance, a per-record
score produced by a GPT-4o-mini judge over the retrieved paths. Each
method's retrieved triplets are assembled into complete multi-hop chains
running from the topic entity to the answer entity, capped at $K = 100$
per record, and presented to the judge together with the query and the GT
evidence triplets. The judge returns a score of $2$ when the retrieved
paths trace a coherent biomedical pathway from topic to answer that
captures the expected mechanism, $1$ when they surface related
neighbours or analogous entities, and $0$ otherwise. Scores are rescaled
to $[0,1]$ and averaged over all records, excluding those where the judge
output fails to parse after three retries.

\section{Per-Depth Traversal Accuracy on 2-hop Queries}
\label{sec:appendix:perdepth2hop}

The main paper reports per-depth hit rate on 3-hop queries in
Figure~\ref{fig:analysis_hit_ratio}, and here we provide the companion
plot for 2-hop queries on the same Llama-3.1-70B backbone
(Figure~\ref{fig:appendix_perdepth_2hop}). The trend mirrors the 3-hop
result, where \sys{} sustains a high hit rate at the answer-bearing
depth across all three query formulations, while path-finding baselines
drop sharply after the topic entity. \sys{} stays ahead of the strongest
baseline at the answer depth under every formulation, consistent with the
pattern observed in the 3-hop figure.

\begin{crblock}
\section{Meta-Path Retrieval Quality}
\label{sec:appendix:metapath}

Path-Bank stores type-level meta-path schemas rather than entity-level
paths. One meta-path such as \emph{drug $\rightarrow$ gene $\rightarrow$
disease} instantiates many entity-level paths, and in \textsc{BioStrat-QA}
the meta-paths are shared across splits while the entity-level paths are
disjoint, so meta-path overlap does not reveal the test path or its
answer. Table~\ref{tab:recall_ctx_relevance} and
Figure~\ref{fig:analysis_hit_ratio} assess the entity-level paths that
traversal produces, and we complement them here by assessing the
meta-paths that guide it. We ask whether \sys{} recovers the reference
meta-path behind each query, and whether the alternatives it retrieves
instead remain traversable to the answer once every exact ground-truth
match is excluded (Table~\ref{tab:appendix_metapath}).

Reference meta-path recovery declines moderately as surface cues are
removed, and the retrieved alternatives remain traversable at a nearly
constant rate across all three information levels. Read together with the
Context Relevance results in Table~\ref{tab:recall_ctx_relevance},
\sys{} supplies usable reasoning schemas rather than reproducing a single
reference path.

\begin{table}[!htbp]
  \centering\small
  \color{black}
  \setlength{\tabcolsep}{4pt}
  \renewcommand{\arraystretch}{1.1}
  \begin{tabularx}{\columnwidth}{@{}l *{6}{>{\centering\arraybackslash}X}@{}}
    \toprule
    \textbf{Level} & \textbf{R@1} & \textbf{R@3} & \textbf{R@5} & \textbf{MRR}
      & \textbf{Trav.}$^{\dagger}$ & \textbf{QA} \\
    \midrule
    \rowcolor{lightgray}
    \multicolumn{7}{c}{\texttt{Llama-3.1-70B}} \\
    \midrule
    Explicit & 66.2 & 79.5 & 82.7 & 0.730 & 55.9 & 71.3 \\
    Implicit & 52.7 & 67.1 & 71.2 & 0.601 & 56.5 & 68.5 \\
    Bare     & 44.8 & 60.1 & 65.7 & 0.527 & 55.0 & 56.5 \\
    \midrule
    \rowcolor{lightgray}
    \multicolumn{7}{c}{\texttt{Qwen-2.5-72B}} \\
    \midrule
    Explicit & 63.2 & 75.8 & 79.0 & 0.697 & 54.4 & 67.4 \\
    Implicit & 51.2 & 65.4 & 69.3 & 0.584 & 55.6 & 65.5 \\
    Bare     & 46.4 & 61.3 & 66.8 & 0.541 & 54.3 & 54.1 \\
    \bottomrule
  \end{tabularx}
  \caption{\textbf{Meta-path retrieval quality on
  \textsc{BioStrat-QA}.} R@$k$ and MRR score recovery and ranking of the
  ground-truth meta-path. Trav.$^{\dagger}$ is the share of retrieved
  meta-paths that can be walked from the topic entity to the answer
  entity on the KG, computed after excluding all exact ground-truth
  matches so that it covers only the retrieved alternatives. QA is
  downstream accuracy.}
  \label{tab:appendix_metapath}
\end{table}

On STaRK-Prime and MedDDx the ground-truth evidence path behind each
query is not released, so the path-level analysis of
Section~\ref{sec:analysis:pathrecover} is not applicable. We instead
measure traversability only, which requires no reference path. Two
differences from the \textsc{BioStrat-QA} setting should be noted. The
template pool underlying both datasets reaches at most two
hops~\citep{wu2024stark}, and the whole retrieved pool is measured here
without excluding exact ground-truth matches, so the numbers are not
directly comparable to Table~\ref{tab:appendix_metapath}. Under this
measurement, traversability remains high in every case, including the
human-generated split written without templates (Table~\ref{tab:appendix_traversability}).

\begin{table}[!htbp]
  \centering\small
  \color{black}
  \setlength{\tabcolsep}{4pt}
  \renewcommand{\arraystretch}{1.1}
  \begin{tabularx}{\columnwidth}{@{}ll *{3}{>{\centering\arraybackslash}X}@{}}
    \toprule
    \textbf{Dataset} & \textbf{Split} & \textbf{$n$}
      & \textbf{Llama-70B} & \textbf{Qwen-72B} \\
    \midrule
    STaRK-Prime & Synthesized & 2{,}801 & 99.0 & 99.3 \\
    STaRK-Prime & Human       & 98      & 86.4 & 89.7 \\
    MedDDx      & Test        & 1{,}769 & 99.2 & 99.4 \\
    \bottomrule
  \end{tabularx}
  \caption{\textbf{Traversability of retrieved meta-paths on
  STaRK-Prime and MedDDx.} Neither dataset releases a reference path, so
  no exact ground-truth match can be excluded and traversability is
  measured over the whole retrieved pool rather than the alternatives
  alone.}
  \label{tab:appendix_traversability}
\end{table}
\end{crblock}

\begin{crblock}
\section{Semantic Robustness of Query-Adaptive Retrieval}
\label{sec:appendix:crossform}

Path-Bank retrieval scores training queries by a mix of dense
sentence-embedding and sparse BM25 similarity
(Eq.~\ref{eq:hybrid_sim}), which raises the question of how much the
retrieved meta-paths depend on the wording of the query rather than the
mechanism behind it. We isolate this by building the Path-Bank from one
query formulation and evaluating on another, holding the underlying
biomedical mechanism fixed. The \emph{same-expression}
setting averages the three matched pairings of explicit, implicit and
bare, and the \emph{different-expression} setting averages the six
mismatched pairings.

Under mismatch, Recall@5 decreases only moderately and QA accuracy stays
well above the strongest baseline
(Table~\ref{tab:appendix_crossform}). Even after every exact
ground-truth match is removed, roughly half or more of the retrieved
meta-paths remain traversable in both settings, so retrieval surfaces
usable alternatives rather than only the schema that generated the
query.

\begin{table}[!htbp]
  \centering\small
  \color{black}
  \setlength{\tabcolsep}{4pt}
  \renewcommand{\arraystretch}{1.1}
  \resizebox{\ifdim\width>\columnwidth \columnwidth\else\width\fi}{!}{
  \begin{tabular}{@{}l ccccc@{}}
    \toprule
    \textbf{Path-Bank / Test} & \textbf{Recall@5} & \textbf{MRR}
      & \textbf{Trav.}$^{\dagger}$ & \textbf{QA Acc.} & \textbf{$\Delta$} \\
    \midrule
    Same-expression      & 73.2 & 0.619 & 55.8 & 65.4 & +22.8 \\
    Different-expression & 68.9 & 0.544 & 49.2 & 61.8 & +19.2 \\
    \bottomrule
  \end{tabular}
  }
  \caption{\textbf{Retrieval and QA under matched and mismatched
  query formulations.} Recall@5 and MRR are computed against the
  ground-truth meta-path. Trav.$^{\dagger}$ excludes all exact
  ground-truth matches, so it covers only the retrieved alternatives.
  $\Delta$ is the gain over the strongest baseline.}
  \label{tab:appendix_crossform}
\end{table}
\end{crblock}

\begin{crblock}
\section{Robustness of Context Relevance}
\label{sec:appendix:crjudge}

LLM-as-judge scoring carries variance across both the choice of judge and
repeated runs, so we verify the Context Relevance results in
Table~\ref{tab:recall_ctx_relevance} against both. We re-score the same
reasoning paths with a second judge, Claude-Haiku-4.5~\citep{anthropic2025haiku45}, alongside the
GPT-4o-mini judge used in the main results, running three trials per
judge and averaging. \sys{} is the best method under both judges and both
backbones, and run-to-run variation stays below half a point
(Table~\ref{tab:appendix_crjudge}).

\begin{table}[!htbp]
  \centering\footnotesize
  \color{black}
  \setlength{\tabcolsep}{2pt}
  \renewcommand{\arraystretch}{1.1}
  \begin{tabularx}{\columnwidth}{@{}l *{4}{>{\centering\arraybackslash}X}@{}}
    \toprule
    \multirow{2}{*}{\textbf{Method}}
      & \multicolumn{2}{c}{\textbf{Llama-3.1-70B}}
      & \multicolumn{2}{c}{\textbf{Qwen-2.5-72B}} \\
    \cmidrule(lr){2-3} \cmidrule(lr){4-5}
      & \makecell{GPT-\\4o-mini} & \makecell{Claude-\\Haiku-4.5}
      & \makecell{GPT-\\4o-mini} & \makecell{Claude-\\Haiku-4.5} \\
    \midrule
    StructGPT & 14.4$\pm$0.12 & 22.3$\pm$0.08 & 29.7$\pm$0.27 & 43.7$\pm$0.04 \\
    ToG       & 16.2$\pm$0.11 & 23.2$\pm$0.21 & 19.0$\pm$0.20 & 25.8$\pm$0.42 \\
    ToG-2     & \underline{37.3$\pm$0.24} & 45.7$\pm$0.26
              & \underline{39.8$\pm$0.16} & 47.7$\pm$0.12 \\
    PoG       & 33.0$\pm$0.32 & \underline{47.4$\pm$0.09}
              & 33.5$\pm$0.17 & \underline{49.2$\pm$0.18} \\
    \rowcolor[HTML]{F4EBFF}
    \textbf{\sys{}} & \textbf{46.7$\pm$0.13} & \textbf{59.6$\pm$0.22}
              & \textbf{47.1$\pm$0.12} & \textbf{60.0$\pm$0.09} \\
    \bottomrule
  \end{tabularx}
  \caption{\textbf{Context Relevance under two LLM judges.}
  Values are mean $\pm$ std over three trials, $\times$100. Per-column
  best is \textbf{bold} and second-best is \underline{underlined}.}
  \label{tab:appendix_crjudge}
\end{table}
\end{crblock}

\section{Ablation Study on Path-Bank \& Meta-Path}
\label{sec:appendix:ablation}

We ablate \sys{}'s query-adaptive meta-path guidance on large backbones,
Llama-3.1-70B and Qwen-2.5-72B, comparing three variants of the
path-finding stage. \textbf{\sys{}} runs the full pipeline with
query-adaptive meta-paths retrieved from the pre-mined Path-Bank.
\textbf{w/o Path-Bank} has the LLM plan meta-paths at inference rather
than retrieving them from the pre-mined Path-Bank, and non-traversable
plans fall back to free BFS. \textbf{w/o meta-path} routes every query
through free BFS with no meta-path guidance. All variants share the same
answer-generation stage, so any gap isolates the effect of
query-adaptive meta-path guidance.

\begin{table*}[!htbp]
  \centering\footnotesize
  \setlength{\tabcolsep}{4pt}
  \setlength{\dashlinedash}{0.5pt}
  \setlength{\dashlinegap}{1.5pt}
  \renewcommand{\arraystretch}{1.05}
  \resizebox{\textwidth}{!}{
  \begin{tabular}{@{}l cccc cccc cccc c@{}}
    \toprule
    \multirow{2}{*}{\textbf{Method}} & \multicolumn{4}{c}{\textbf{Explicit}} & \multicolumn{4}{c}{\textbf{Implicit}} & \multicolumn{4}{c}{\textbf{Bare}} & \multirow{2}{*}{\textbf{Overall}} \\
    \cmidrule(lr){2-5} \cmidrule(lr){6-9} \cmidrule(lr){10-13}
    & 1-hop & 2-hop & 3-hop & Avg & 1-hop & 2-hop & 3-hop & Avg & 1-hop & 2-hop & 3-hop & Avg & \\
    \midrule
    \rowcolor{lightgray}
    \multicolumn{14}{c}{\texttt{Llama-3.1-70B}} \\
    \midrule
    w/o meta-path  & 81.2 & 43.2 & 35.0 & 55.8 & 77.1 & 41.3 & 38.2 & 54.0 & \underline{68.4} & \underline{34.5} & \underline{34.1} & \underline{47.0} & 52.3 \\
    w/o Path-Bank  & \underline{84.0} & \underline{57.1} & \underline{41.5} & \underline{64.2} & \underline{79.0} & \underline{51.0} & \underline{39.2} & \underline{59.2} & 64.5 & 32.0 & 29.9 & 43.7 & \underline{55.7} \\
    \rowcolor[HTML]{F4EBFF} \textbf{\sys{}}
               & \textbf{87.9} & \textbf{61.3} & \textbf{62.2} & \textbf{71.3} & \textbf{83.1} & \textbf{62.5} & \textbf{53.9} & \textbf{68.5} & \textbf{69.8} & \textbf{50.5} & \textbf{44.2} & \textbf{56.5} & \textbf{65.5} \\
    \midrule
    \rowcolor{lightgray}
    \multicolumn{14}{c}{\texttt{Qwen-2.5-72B}} \\
    \midrule
    w/o meta-path  & 78.3 & 36.0 & 33.6 & 51.2 & 73.9 & 37.9 & \underline{36.9} & 51.1 & 66.6 & 28.2 & 32.3 & 43.2 & 48.5 \\
    w/o Path-Bank  & \underline{81.7} & \underline{50.9} & \underline{41.0} & \underline{60.5} & \underline{81.7} & \underline{45.5} & 34.6 & \underline{56.9} & \underline{69.1} & \underline{28.6} & \underline{34.6} & \underline{44.7} & \underline{54.0} \\
    \rowcolor[HTML]{F4EBFF} \textbf{\sys{}}
               & \textbf{84.9} & \textbf{58.3} & \textbf{53.9} & \textbf{67.4} & \textbf{83.3} & \textbf{59.2} & \textbf{44.7} & \textbf{65.5} & \textbf{71.9} & \textbf{44.8} & \textbf{41.0} & \textbf{54.1} & \textbf{62.3} \\
    \bottomrule
  \end{tabular}
  }
  \caption{\textbf{Ablation on Path-Bank and Meta-Path, \textsc{BioStrat-QA}, large backbones.}
  Exact-match (\%) per cell, where \textbf{Overall} averages all nine
  cells. Per-column best is \textbf{bold} and second-best is
  \underline{underlined} within each backbone block.}
  \label{tab:appendix_ablation_biostrat}
\end{table*}

On \textsc{BioStrat-QA}, \sys{} leads both ablations on every cell, and
the gap widens as reasoning paths extend
(Table~\ref{tab:appendix_ablation_biostrat}). On 1-hop queries, where a single edge separates topic from
answer, the two ablations stay close to \sys{} across all three
information levels. The gap opens up on 2-hop and 3-hop queries, where an
unguided search has to commit to intermediate entities before reaching
the answer. Such a pattern indicates that query-adaptive meta-path
guidance carries an increasing share of the answer-generation accuracy as
path-finding moves into deeper regions of the biomedical KG.

\begin{table}[!htbp]
  \centering\footnotesize
  \setlength{\tabcolsep}{3pt}
  \setlength{\dashlinedash}{0.5pt}
  \setlength{\dashlinegap}{1.5pt}
  \renewcommand{\arraystretch}{1.05}
  \resizebox{\columnwidth}{!}{
  \begin{tabular}{@{}l cc cccc@{}}
    \toprule
    \multirow{2}{*}{\textbf{Method}} & \multicolumn{2}{c}{\textbf{STaRK-Prime}} & \multicolumn{4}{c}{\textbf{MedDDx}} \\
    \cmidrule(lr){2-3} \cmidrule(lr){4-7}
    & Synthesized & \begin{tabular}{@{}c@{}}Human-\\generated\end{tabular} & Basic & Inter. & Expert & All \\
    \midrule
    \rowcolor{lightgray}
    \multicolumn{7}{c}{\texttt{Llama-3.1-70B}} \\
    \midrule
    w/o meta-path  & 37.0 & 38.8 & \underline{62.0} & 50.9 & \underline{51.1} & 52.5 \\
    w/o Path-Bank  & \underline{41.2} & \underline{39.8} & 59.6 & \underline{53.0} & 50.3 & \underline{53.2} \\
    \rowcolor[HTML]{F4EBFF} \textbf{\sys{}}
            & \textbf{46.9} & \textbf{43.9} & \textbf{65.3} & \textbf{55.1} & \textbf{52.4} & \textbf{55.8} \\
    \midrule
    \rowcolor{lightgray}
    \multicolumn{7}{c}{\texttt{Qwen-2.5-72B}} \\
    \midrule
    w/o meta-path  & 39.2 & \underline{38.8} & 58.4 & \underline{54.4} & 46.0 & \underline{52.6} \\
    w/o Path-Bank  & \underline{43.2} & 35.7 & \underline{60.4} & 52.5 & \underline{47.8} & 52.3 \\
    \rowcolor[HTML]{F4EBFF} \textbf{\sys{}}
            & \textbf{46.7} & \textbf{39.8} & \textbf{62.0} & \textbf{54.5} & \textbf{51.8} & \textbf{54.8} \\
    \bottomrule
  \end{tabular}
  }
  \caption{\textbf{Ablation on Path-Bank and Meta-Path, STaRK-Prime and
  MedDDx, large backbones.} Exact-match (\%) on STaRK-Prime and
  multiple-choice accuracy (\%) on MedDDx. Per-column best is
  \textbf{bold} and second-best is \underline{underlined} within each
  backbone block.}
  \label{tab:appendix_ablation_starkmed}
\end{table}

We also run the same ablation on the external datasets STaRK-Prime and
MedDDx (Table~\ref{tab:appendix_ablation_starkmed}). On STaRK-Prime,
\sys{} stays ahead of both ablations on the synthesized and
human-generated splits across both backbones. On MedDDx, whose tiers
order distractors by semantic closeness to the gold answer, \sys{} stays
ahead of both ablations on every tier under both backbones, so the
guidance holds up even when the answer options are hard to tell apart.

\begin{crblock}
\section{Hyperparameter Settings}
\label{sec:appendix:hyperparams}

The hyperparameters used for Path-Bank construction and inference are
listed in Table~\ref{tab:appendix_hyperparams}.

\begin{table}[!htbp]
  \centering\small
  \color{black}
  \setlength{\tabcolsep}{5pt}
  \renewcommand{\arraystretch}{1.15}
  \begin{tabularx}{\columnwidth}{@{}cl >{\raggedright\arraybackslash}X@{}}
    \toprule
    \textbf{Symbol} & \textbf{Value} & \textbf{Description} \\
    \midrule
    $\beta$  & 0.3  & Trade-off between relation and endpoint similarity in
                      edge weighting (Eq.~\ref{eq:edge_weight}) \\
    $\eta$   & 0.85 & Teleport probability of the PPR walks
                      (Eqs.~\ref{eq:ppr_fwd},~\ref{eq:ppr_bwd}) \\
    $K$      & 150  & Corridor subgraph size in nodes
                      (Eq.~\ref{eq:corridor}) \\
    $\alpha$ & 0.5  & Trade-off between edge weight and corridor score in the
                      path cost (Eq.~\ref{eq:yen_cost}) \\
    $n$      & 5    & Paths extracted per hop length by Yen's algorithm \\
    $\lambda$ & 0.5 & Trade-off between sentence-embedding and BM25 similarity
                      in retrieval (Eq.~\ref{eq:hybrid_sim}) \\
    $k$      & 5    & Top-ranked training queries retained as candidates \\
    \bottomrule
  \end{tabularx}
  \caption{\textbf{Hyperparameters for Path-Bank
  construction and inference.} Values are shared across datasets and
  backbones.}
  \label{tab:appendix_hyperparams}
\end{table}
\end{crblock}

\begin{crblock}
\section{Efficiency Analysis}
\label{sec:appendix:efficiency}

\subsection{Offline Path-Bank Construction}
\label{sec:appendix:efficiency:offline}

Path-Bank is built once, offline. Its costly stage is Yen's
$k$-shortest-path search, which we run inside the query-specific
corridor produced by edge weighting and bidirectional PPR instead of
over the full KG. To quantify the saving, we profile 360
\textsc{BioStrat-QA} queries sampled evenly across the three
formulations and reasoning depths, varying the KG from 30K to 129K nodes
and timing each stage. The corridor itself comes from edge weighting and
PPR, so the two settings differ only in where Yen's search runs.

Edge weighting and PPR stay inexpensive as the KG grows, whereas Yen's
search over the raw KG grows sharply in both runtime and memory
(Tables~\ref{tab:appendix_offline_time}
and~\ref{tab:appendix_offline_mem}). Confining it to the corridor keeps
both nearly flat. Over all 7{,}473 training queries, corridor-restricted
search finishes in under ten CPU-minutes, against dozens of CPU-hours
for the raw KG.

\begin{table}[!htbp]
  \centering\small
  \color{black}
  \setlength{\tabcolsep}{4pt}
  \renewcommand{\arraystretch}{1.1}
  \resizebox{\ifdim\width>\columnwidth \columnwidth\else\width\fi}{!}{
  \begin{tabular}{@{}ll cc cc@{}}
    \toprule
    \multirow{2}{*}{\textbf{Nodes}} & \multirow{2}{*}{\textbf{Edges}}
      & \multirow{2}{*}{\textbf{Edge wt.}} & \multirow{2}{*}{\textbf{PPR}}
      & \multicolumn{2}{c}{\textbf{Yen's search}} \\
    \cmidrule(lr){5-6}
    & & & & Raw KG & Corridor \\
    \midrule
    30K  & 552K  & 94\,ms  & 65\,ms  & 1.52\,s  & 77\,ms \\
    60K  & 1.90M & 323\,ms & 117\,ms & 4.52\,s  & 78\,ms \\
    90K  & 4.00M & 662\,ms & 201\,ms & 9.95\,s  & 87\,ms \\
    129K & 8.10M & 1.34\,s & 427\,ms & 19.57\,s & 69\,ms \\
    \bottomrule
  \end{tabular}
  }
  \caption{\textbf{Runtime of Path-Bank construction by KG
  size.} Edge weighting and PPR are shared by both settings.}
  \label{tab:appendix_offline_time}
\end{table}

\begin{table}[!htbp]
  \centering\small
  \color{black}
  \setlength{\tabcolsep}{4pt}
  \renewcommand{\arraystretch}{1.1}
  \resizebox{\ifdim\width>\columnwidth \columnwidth\else\width\fi}{!}{
  \begin{tabular}{@{}ll cc cc@{}}
    \toprule
    \multirow{2}{*}{\textbf{Nodes}} & \multirow{2}{*}{\textbf{Edges}}
      & \multirow{2}{*}{\textbf{Edge wt.}} & \multirow{2}{*}{\textbf{PPR}}
      & \multicolumn{2}{c}{\textbf{Yen's search}} \\
    \cmidrule(lr){5-6}
    & & & & Raw KG & Corridor \\
    \midrule
    30K  & 552K  & 44\,MB  & 2.6\,MB  & 48\,MB          & 0.47\,MB \\
    60K  & 1.90M & 151\,MB & 5.3\,MB  & 166\,MB         & 0.48\,MB \\
    90K  & 4.00M & 318\,MB & 7.9\,MB  & 351\,MB         & 0.47\,MB \\
    129K & 8.10M & 640\,MB & 11.4\,MB & $\sim$711\,MB   & $\sim$0.40\,MB \\
    \bottomrule
  \end{tabular}
  }
  \caption{\textbf{Peak memory of Path-Bank construction by
  KG size.} The corridor variant stays under 0.5\,MB,
  since Yen's search never leaves the extracted subgraph.}
  \label{tab:appendix_offline_mem}
\end{table}

\subsection{Online Path-Finding}
\label{sec:appendix:efficiency:online}

At inference the cost is dominated by the LLM scoring that selects the
next-hop frontier, a stage shared with the other path-finding
baselines, while Path-Bank matching is negligible
(Table~\ref{tab:appendix_online_stage}).
Table~\ref{tab:appendix_online_e2e} compares all path-finding methods
under the same backbone and search budget, reporting cost together with
accuracy.

\sys{} issues fewer LLM calls and far fewer tokens than ToG and PoG
while running faster and reaching higher accuracy. StructGPT and ToG-2
are cheaper still, but explore far less and land well below \sys{} in EM.
Restricting traversal to the relation and node types of the retrieved
meta-paths, and stopping at the required reasoning depth, keeps the
search bounded.

\begin{table}[!htbp]
  \centering\small
  \color{black}
  \setlength{\tabcolsep}{4pt}
  \renewcommand{\arraystretch}{1.1}
  \resizebox{\ifdim\width>\columnwidth \columnwidth\else\width\fi}{!}{
  \begin{tabular}{@{}lll@{}}
    \toprule
    \textbf{Stage} & \textbf{Runtime / query} & \textbf{Memory} \\
    \midrule
    Path-Bank matching   & 0.1\,ms \ \ (0.0002\%) & 524\,MB CPU, 2.6\,GB GPU \\
    LLM next-hop scoring & 42.89\,s \ (95.2\%)    & 0.72\,MB per query \\
    Other operations     & 2.14\,s \ \ (4.8\%)    & included in peak \\
    \midrule
    Total                & 45.03\,s \ (100\%)     & \\
    \bottomrule
  \end{tabular}
  }
  \caption{\textbf{Per-stage online cost of \sys{}.} Memory
  for Path-Bank matching is a one-time footprint rather than a per-query
  cost.}
  \label{tab:appendix_online_stage}
\end{table}

\begin{table}[!htbp]
  \centering\small
  \color{black}
  \setlength{\tabcolsep}{4pt}
  \renewcommand{\arraystretch}{1.1}
  \resizebox{\ifdim\width>\columnwidth \columnwidth\else\width\fi}{!}{
  \begin{tabular}{@{}l c rr r r@{}}
    \toprule
    \textbf{Method} & \textbf{Latency (s)} & \textbf{Calls} & \textbf{Tokens}
      & \textbf{Mem.\,(MB)} & \textbf{EM} \\
    & mean / max & & & & \\
    \midrule
    StructGPT & 7.7 / 15.7    & 5.5   & 2{,}175   & 671     & 30.4 \\
    ToG-2     & 13.5 / 34.4   & 5.8   & 2{,}565   & 4{,}895 & 35.2 \\
    ToG       & 152.0 / 803.9 & 17.7  & 13{,}285  & 2{,}429 & 40.0 \\
    PoG       & 64.8 / 507.7  & 289.3 & 155{,}150 & 3{,}565 & 39.9 \\
    \rowcolor[HTML]{F4EBFF}
    \textbf{\sys{}} & 45.0 / 164.5 & 11.2 & 7{,}036 & 4{,}186 & \textbf{65.5} \\
    \bottomrule
  \end{tabular}
  }
  \caption{\textbf{End-to-end online cost and accuracy.}
  All methods share the same search budget on Llama-3.1-70B. Calls and
  tokens are per query.}
  \label{tab:appendix_online_e2e}
\end{table}
\end{crblock}

\begin{crblock}
\section{Statistical Significance}
\label{sec:appendix:significance}

We assess significance with a paired bootstrap over the main results in
Tables~\ref{tab:main_results_biotog} and~\ref{tab:stark_medddx_results},
resampling the per-query correct or incorrect outcomes 10{,}000 times
and taking the 2.5th and 97.5th percentiles of the accuracy difference.
Comparing \sys{} against each baseline in every setting and backbone
yields 270 pairwise comparisons, of which 255 are significant.
Table~\ref{tab:appendix_significance} reports the interval against the
strongest baseline in each setting.

The remaining cases sit in the smallest subsets, such as the STaRK-Prime
human-written split and MedDDx Basic, where the limited sample size
widens the interval. The point estimate stays positive in every one of
these settings.

\begin{table*}[!t]
  \centering\footnotesize
  \color{black}
  \setlength{\tabcolsep}{4pt}
  \renewcommand{\arraystretch}{1.02}
  \begin{tabularx}{\textwidth}{@{}l >{\centering\arraybackslash}X l >{\centering\arraybackslash}X l@{}}
    \toprule
    \multirow{2}{*}{\textbf{Setting}}
      & \multicolumn{2}{c}{\textbf{Llama-3.1-70B}}
      & \multicolumn{2}{c}{\textbf{Qwen-2.5-72B}} \\
    \cmidrule(lr){2-3} \cmidrule(lr){4-5}
      & $\Delta$ [95\% CI] & Baseline & $\Delta$ [95\% CI] & Baseline \\
    \midrule
    \rowcolor{lightgray}
    \multicolumn{5}{c}{\textsc{BioStrat-QA}} \\
    \midrule
    Explicit 1-hop & +5.0 [+0.9, +9.2]     & PoG       & +8.9 [+4.6, +13.3]  & PoG \\
    Explicit 2-hop & +22.9 [+18.1, +27.6]  & CoK       & +24.4 [+19.6, +29.1] & CoK \\
    Explicit 3-hop & +25.3 [+17.5, +33.2]  & StructGPT & +19.8 [+12.0, +27.6] & StructGPT \\
    Implicit 1-hop & +5.7 [+0.9, +10.5]    & PoG       & +11.2 [+6.6, +16.0] & PoG \\
    Implicit 2-hop & +22.5 [+17.7, +27.0]  & CoT       & +22.5 [+17.5, +27.4] & CoK \\
    Implicit 3-hop & +13.4 [+6.5, +20.3]   & CoK       & +6.5 [$-$0.5, +13.4] & SC \\
    Bare 1-hop     & +6.9 [+1.6, +11.9]    & PoG       & +12.1 [+7.3, +16.9] & PoG \\
    Bare 2-hop     & +14.1 [+9.5, +18.9]   & CoK       & +12.6 [+7.8, +17.3] & IO \\
    Bare 3-hop     & +4.1 [$-$3.7, +12.0]  & StructGPT & +3.2 [$-$3.2, +9.7] & CoT \\
    \midrule
    \rowcolor{lightgray}
    \multicolumn{5}{c}{STaRK-Prime and MedDDx} \\
    \midrule
    STaRK-Prime (synth.) & +5.5 [+3.4, +7.7]    & PoG       & +13.2 [+11.1, +15.4] & PoG \\
    STaRK-Prime (human)  & +3.1 [$-$7.1, +13.3] & PoG       & +4.1 [$-$6.1, +14.3] & ToG-2 \\
    MedDDx Basic         & +4.9 [$-$0.8, +10.6] & ToG       & +2.9 [$-$3.7, +9.4]  & StructGPT \\
    MedDDx Inter.        & +6.8 [+3.6, +10.1]   & StructGPT & +4.5 [+1.4, +7.6]    & StructGPT \\
    MedDDx Expert        & +4.8 [$-$0.4, +9.9]  & KGARevion & +8.5 [+4.1, +12.8]   & ToG \\
    MedDDx All           & +6.0 [+3.7, +8.4]    & ToG       & +5.7 [+3.3, +8.0]    & StructGPT \\
    \bottomrule
  \end{tabularx}
  \caption{\textbf{Accuracy difference over the strongest
  baseline with 95\% bootstrap intervals.} The Baseline column names
  the strongest baseline in that setting. An interval that excludes zero
  is a significant gain.}
  \label{tab:appendix_significance}
\end{table*}
\end{crblock}

\begin{crblock}
\section{Generalization to an External Biomedical KG}
\label{sec:appendix:hetionet}

We examine whether \sys{}'s path-finding scheme remains effective on a
knowledge graph other than PrimeKG. We build a Hetionet-based KGQA
dataset following the \textsc{BioStrat-QA} construction procedure, with
explicit, implicit and bare variants of every query. Hetionet
\citep{himmelstein2017systematic} differs from PrimeKG in scale, degree
distribution, entity and relation types, and available textual metadata.
We compare against the LLM-only floor and the path-finding baselines,
with all methods running on Llama-3.1-70B under the same protocol as the
main experiments.

\sys{} is the strongest method at every information level
(Table~\ref{tab:appendix_hetionet}), and its margin over the
path-finding baselines widens as intermediate cues disappear.
Query-adaptive meta-path retrieval from Path-Bank therefore transfers as
a path-finding scheme to a knowledge graph built independently of
PrimeKG.

\begin{table}[!htbp]
  \centering\small
  \color{black}
  \setlength{\tabcolsep}{4pt}
  \setlength{\dashlinedash}{0.5pt}
  \setlength{\dashlinegap}{1.5pt}
  \renewcommand{\arraystretch}{1.1}
  \begin{tabularx}{\columnwidth}{@{}l *{3}{>{\centering\arraybackslash}X}@{}}
    \toprule
    \textbf{Method} & \textbf{Explicit} & \textbf{Implicit} & \textbf{Bare} \\
    \midrule
    IO        & 39.0 & 36.9 & 27.7 \\
    CoT       & 39.7 & 39.0 & 32.6 \\
    SC        & 44.0 & 40.0 & 29.8 \\

    \noalign{\vskip 2pt}    \cdashline{1-4}    \noalign{\vskip 2pt}

    StructGPT & 60.3 & \underline{48.9} & \underline{41.1} \\
    ToG       & \underline{61.7} & 41.1 & 38.3 \\
    PoG       & 53.2 & 39.7 & 37.6 \\
    \rowcolor[HTML]{F4EBFF}
    \textbf{\sys{}} & \textbf{78.0} & \textbf{68.8} & \textbf{67.4} \\
    \bottomrule
  \end{tabularx}
  \caption{\textbf{QA accuracy on Hetionet KGQA.} The three
  information levels are generated from the same reference path, so they
  differ only in how much intermediate reasoning the query exposes.
  Per-column best is \textbf{bold} and second-best is
  \underline{underlined}.}
  \label{tab:appendix_hetionet}
\end{table}
\end{crblock}

\begin{crblock}
\section{LLM-Inferred Control Signals}
\label{sec:appendix:control}

\sys{} filters the retrieved meta-paths by an LLM-inferred hop length and
answer type (Section~\ref{sec:method:pipeline}). We check how reliably
the two are predicted, and how much the framework depends on getting them
right, by replacing each prediction with its ground-truth value.

Answer-type prediction stays high regardless of information level, and
hop prediction assigns the correct depth to the majority of queries at
every level, with accuracy decreasing from explicit to bare as the query
retains fewer intermediate cues to infer depth from
(Table~\ref{tab:appendix_control_acc}). Replacing both predictions with
their ground-truth values shifts performance by only a few points in
either direction (Table~\ref{tab:appendix_control_oracle}), so the LLM
inference of these control signals is not a bottleneck for the
framework.

\begin{table}[!htbp]
  \centering\small
  \color{black}
  \setlength{\tabcolsep}{3pt}
  \renewcommand{\arraystretch}{1.1}
  \begin{tabularx}{\columnwidth}{@{}l *{4}{>{\centering\arraybackslash}X}@{}}
    \toprule
    \multirow{2}{*}{\textbf{Query level}}
      & \multicolumn{2}{c}{\textbf{Llama-3.1-70B}}
      & \multicolumn{2}{c}{\textbf{Qwen-2.5-72B}} \\
    \cmidrule(lr){2-3} \cmidrule(lr){4-5}
      & Hop & Ans Type & Hop & Ans Type \\
    \midrule
    Explicit & 90.1 & 97.1 & 88.9 & 93.6 \\
    Implicit & 80.6 & 90.1 & 80.7 & 88.7 \\
    Bare     & 68.6 & 91.8 & 70.6 & 92.5 \\
    \midrule
    All      & 79.8 & 93.0 & 80.0 & 91.6 \\
    \bottomrule
  \end{tabularx}
  \caption{\textbf{Control-signal prediction accuracy
  (\%).} Hop and Ans Type are scored against the ground-truth reasoning
  depth and answer entity type.}
  \label{tab:appendix_control_acc}
\end{table}

\begin{table}[!htbp]
  \centering\small
  \color{black}
  \setlength{\tabcolsep}{3pt}
  \renewcommand{\arraystretch}{1.1}
  \begin{tabularx}{\columnwidth}{@{}l *{6}{>{\centering\arraybackslash}X}@{}}
    \toprule
    \multirow{2}{*}{\textbf{Formulation}}
      & \multicolumn{2}{c}{\textbf{1-hop}}
      & \multicolumn{2}{c}{\textbf{2-hop}}
      & \multicolumn{2}{c}{\textbf{3-hop}} \\
    \cmidrule(lr){2-3} \cmidrule(lr){4-5} \cmidrule(lr){6-7}
      & Pred. & Oracle & Pred. & Oracle & Pred. & Oracle \\
    \midrule
    Explicit & 87.9 & 88.3 & 61.3 & 58.9 & 62.2 & 61.3 \\
    Implicit & 83.1 & 87.0 & 62.5 & 62.5 & 53.9 & 51.2 \\
    Bare     & 69.8 & 76.2 & 50.5 & 52.0 & 44.2 & 47.0 \\
    \bottomrule
  \end{tabularx}
  \caption{\textbf{Oracle substitution on control signals
  (\%).} Oracle replaces both the predicted hop length and answer type
  with ground-truth values, on Llama-3.1-70B.}
  \label{tab:appendix_control_oracle}
\end{table}
\end{crblock}

\section{Case Studies}
\label{sec:appendix:cases}
To illustrate how \sys{} converts query-adaptive meta-path guidance into
a correct answer, we walk through individual cases on Llama-3.1-70B,
comparing \textbf{CoT}, \textbf{ToG}, and \textbf{\sys{}} on the same
evidence path under all three query formulations.

\begin{crblock}
We first assess whether the reasoning each method produces holds up to
domain scrutiny. A blinded human evaluation covers 10 sampled questions
spanning Llama and Qwen outputs, with anonymized outputs from CoT, ToG
and \sys{} presented in random order. ToG and \sys{} reason over
triplets retrieved from the graph, whereas CoT relies on the backbone
alone. Nine biomedical experts, each with over seven years of
experience, who received a compensation of \$5 upon
completion of the task, independently rated every output on two 1--5 scales.
\textbf{Validity} asks whether the reasoning steps are biologically
valid, and \textbf{Relevance} asks whether the reasoning traces a pathway
that contributes to answering the question. The scales were rated
independently, since biologically true reasoning may still fail to
address the question. The instructions, shown verbatim
below, stated the purpose of the study and the two rating scales.
Ratings were collected solely for aggregate statistics reported in this
paper, and no personal data was collected.
\end{crblock}

\begin{tcolorbox}[colback=gray!4,colframe=gray!40,boxrule=0.5pt,
                  arc=2pt,left=4pt,right=4pt,top=3pt,bottom=3pt,
                  fontupper=\footnotesize\ttfamily,breakable]
\textbf{Introduction}\\[4pt]
\textbf{PURPOSE}\\
This study evaluates how well a knowledge-graph QA system reasons over
biomedical questions. Each answer is produced by tracing a reasoning
path, a chain of entities connected by typed relations (for example:
disease -> gene -> drug). Automatic metrics only check whether a path
exactly matches a reference path, so they cannot tell whether a
differing path is still biologically sensible and still relevant to the
question. We ask you, as a domain expert, to judge exactly that.\\[4pt]
\textbf{WHAT YOU WILL DO}\\
For each of the 10 cases you will see one biomedical question and three
reasoning outputs labeled A, B, and C (presented in random order). Some
reasoning outputs are very long, so we show a curated view focusing on
the main steps. Rate each output on the two scales below.\\[4pt]
\textbf{SCALES (1-5, applied to A, B, and C separately)}\\
(1) Biomedical validity: is the reasoning biomedically meaningful,
i.e., are the stated reasoning steps biologically valid?\\
\hspace*{2em}5 = fully valid \quad 3 = partially valid \quad 1 = not
valid / unrelated\\
(2) Pathway relevance: does the reasoning trace a relevant pathway that
contributes to answering this specific question?\\
\hspace*{2em}5 = directly relevant / decisive \quad 3 = partially
relevant \quad 1 = irrelevant\\[4pt]
A reasoning can score high on one scale and low on the other (e.g.,
biologically true statements that do not actually address the
question). Please rate the two independently. There are no right or
wrong responses; judge based on your expertise.
\end{tcolorbox}

\begin{crblock}
As shown in Table~\ref{tab:appendix_humaneval}, \sys{} receives the highest ratings on both, so the paths it retrieves support
reasoning that is biologically sound and pertinent to the question. ToG falls
below CoT on each, which may indicate that retrieved paths degrade
reasoning quality when they diverge from the mechanism the question
concerns.
\end{crblock}

\begin{crblock}
\begin{table}[!htbp]
  \centering\small
  \color{black}
  \setlength{\tabcolsep}{4pt}
  \renewcommand{\arraystretch}{1.1}
  \begin{tabularx}{\columnwidth}{@{}l *{2}{>{\centering\arraybackslash}X}@{}}
    \toprule
    \textbf{Method} & \textbf{Validity} & \textbf{Relevance} \\
    \midrule
    CoT & 3.70 $\pm$ 1.04 & 3.22 $\pm$ 1.03 \\
    ToG & 2.98 $\pm$ 0.99 & 2.63 $\pm$ 1.26 \\
    \rowcolor[HTML]{F4EBFF}
    \textbf{\sys{}} & \textbf{4.28 $\pm$ 1.05} & \textbf{4.32 $\pm$ 1.08} \\
    \bottomrule
  \end{tabularx}
  \caption{\textbf{Blinded expert ratings of reasoning
  paths.} Nine biomedical experts scored 10 questions on two independent
  1--5 scales. Values are mean $\pm$ std across raters and questions.}
  \label{tab:appendix_humaneval}
\end{table}
\end{crblock}

We further provide the reasoning traces of each method as per-case
tables. Each case is built around one ground-truth path, and for every
method we report the intermediate triplets visited during path-finding
together with the reasoning leading to the final answer. Excerpts are
trimmed with $\ldots$ ellipses around the key intermediate entities. The
trailing ``\textbf{the answer is \{$\cdot$\}}'' phrase is
\textcolor{blue}{\textbf{blue}} when the final answer matches the ground
truth and \textcolor{red}{\textbf{red}} when it does not, and cases
without a parseable answer show \textcolor{red}{\textbf{$\times$}}.
The advantage of \sys{} over path-finding baselines grows as the query
exposes less of the mechanism
(Tables~\ref{tab:appendix_case_3940}, \ref{tab:appendix_case_3786},
\ref{tab:appendix_case_4416},
\ref{tab:appendix_case_4423}).
Explicit queries name the biomedical mechanism in enough detail that an
LLM can often answer them from prior knowledge, so a baseline may reach
the correct answer even when its traversal misses the ground-truth
triplets. ToG answers the explicit query correctly in every case below,
in most of them while recovering few or none of the reference triplets.
Implicit and bare queries leave the mechanism unstated, so a method that
drifts away from the reference path has little left to recover the answer
from, and ToG tends to settle on neighboring entities instead. \sys{}
traces the ground-truth path under all three formulations.

\providecommand{\Acorrect}[1]{\textcolor{blue}{\textbf{the answer is \{#1\}}}}
\providecommand{\Awrong}[1]{\textcolor{red}{\textbf{the answer is \{#1\}}}}
\providecommand{\Anone}{\textcolor{red}{\textbf{\large $\times$}}}
\providecommand{\Cmark}{{\color{black!60}\checkmark}}
\providecommand{\Xmark}{{\color{red}\large $\times$}}

\begin{table*}[p]
\centering\footnotesize
\setlength{\tabcolsep}{6pt}
\renewcommand{\arraystretch}{1.3}
\newcommand{\Apc}{\textit{Anterior plagiocephaly}}
\newcommand{\Cdc}{\textit{CDC45}}
\newcommand{\Dsb}{\textit{double-strand break repair via break-induced replication}}

\begin{tabularx}{\textwidth}{@{}>{\bfseries}p{0.07\textwidth} X X X@{}}
\toprule
 & \normalfont\textbf{Explicit} & \normalfont\textbf{Implicit} & \normalfont\textbf{Bare} \\
\midrule
Query &
\normalfont\textit{\small Anterior plagiocephaly is associated with CDC45; with which DNA repair biological process does CDC45 interact that involves restarting DNA synthesis from a break site?} &
\normalfont\textit{\small In anterior plagiocephaly, alterations in a highly conserved factor that initiates genome duplication suggest disruption of what DNA repair process that restores broken chromosomes by restarting DNA synthesis from a break site?} &
\normalfont\textit{\small Which DNA damage repair process is linked to Anterior plagiocephaly?} \\
\midrule
\multicolumn{4}{@{}p{\textwidth}@{}}{\normalfont\textbf{Ground-truth path}: \Apc{} $\xrightarrow{\text{associated with}}$ \Cdc{} $\xrightarrow{\text{interacts with}}$ \Dsb{}} \\
\midrule
\addlinespace[2pt]
\multicolumn{4}{@{}l}{\normalfont\itshape CoT} \\
\addlinespace[2pt]
Reasoning &
\normalfont $\ldots$ \textbf{break-induced replication} is a type of DNA repair mechanism for double-strand breaks $\ldots$ \newline \Acorrect{break-induced replication} &
\normalfont $\ldots$ cycles between the \emph{ATR/Chk1} and \emph{FoxM1} pathways without converging on a final answer $\ldots$ \newline \Anone &
\normalfont $\ldots$ no specific DNA damage repair process found to be linked to \emph{anterior plagiocephaly} $\ldots$ \newline \Awrong{None} \\
\midrule
\addlinespace[2pt]
\multicolumn{4}{@{}l}{\normalfont\itshape ToG} \\
\addlinespace[2pt]
Path-finding &
\normalfont $d_1$: \Apc{} $\to$ \Cdc{} \,\Cmark \newline
$d_2$: \Cdc{} $\to$ \Dsb{} \,\Cmark &
\normalfont $d_1$: \Apc{} $\to$ \Cdc{} \,\Cmark \newline
$d_2$: \Cdc{} $\to$ \emph{Activation of ATR} \,\Xmark &
\normalfont $d_1$: \Apc{} $\to$ \Cdc{} \,\Cmark \newline
$d_2$: \Cdc{} $\to$ \emph{cerebral cortex} \,\Xmark \\
Reasoning &
\normalfont $\ldots$ \textbf{CDC45} interacts with \textbf{double-strand break repair via break-induced replication} $\ldots$ \newline \Acorrect{double-strand break repair via break-induced replication} &
\normalfont $\ldots$ \textbf{CDC45} interacts with \emph{Activation of ATR}, a parent of \emph{G2/M Checkpoints} $\ldots$ \newline \Awrong{G2/M Checkpoints} &
\normalfont $\ldots$ \textbf{CDC45} interacts with \emph{Activation of ATR}, a parent of \emph{G2/M Checkpoints} $\ldots$ \newline \Awrong{G2/M Checkpoints} \\
\midrule
\addlinespace[2pt]
\multicolumn{4}{@{}l}{\normalfont\itshape \sys{}} \\
\addlinespace[2pt]
Path-finding &
\normalfont $d_1$: \Apc{} $\to$ \Cdc{} \,\Cmark \newline
$d_2$: \Cdc{} $\to$ \Dsb{} \,\Cmark &
\normalfont $d_1$: \Apc{} $\to$ \Cdc{} \,\Cmark \newline
$d_2$: \Cdc{} $\to$ \Dsb{} \,\Cmark &
\normalfont $d_1$: \Apc{} $\to$ \Cdc{} \,\Cmark \newline
$d_2$: \Cdc{} $\to$ \Dsb{} \,\Cmark \\
Reasoning &
\normalfont $\ldots$ \textbf{Anterior plagiocephaly} is associated with \textbf{CDC45}, which interacts with \textbf{double-strand break repair via break-induced replication} $\ldots$ \newline \Acorrect{double-strand break repair via break-induced replication} &
\normalfont $\ldots$ \textbf{anterior plagiocephaly} $\to$ \textbf{CDC45} $\to$ \textbf{double-strand break repair via break-induced replication}, which restarts DNA synthesis from the break site $\ldots$ \newline \Acorrect{double-strand break repair via break-induced replication} &
\normalfont $\ldots$ \textbf{Anterior plagiocephaly} $\to$ \textbf{CDC45} $\to$ \textbf{double-strand break repair via break-induced replication}, a DNA damage repair process $\ldots$ \newline \Acorrect{double-strand break repair via break-induced replication} \\
\bottomrule
\end{tabularx}
\caption{\textbf{Case study, 2-hop, data\_id=3940.}
CoT does not perform path-finding. ToG recovers the correct second-hop
triplet only under the explicit query and drifts to neighboring processes
such as Activation of ATR and cerebral cortex once the intermediate cue
is removed, leading to incorrect final answers. \sys{} reaches the
ground-truth triplet at both depths under all three formulations.}
\label{tab:appendix_case_3940}
\end{table*}

\begin{table*}[p]
\centering\footnotesize
\setlength{\tabcolsep}{6pt}
\renewcommand{\arraystretch}{1.3}
\newcommand{\Bpm}{\textit{basal plasma membrane}}
\newcommand{\Tf}{\textit{TF}}
\newcommand{\Bs}{\textit{Bismuth subsalicylate}}
\begin{tabularx}{\textwidth}{@{}>{\bfseries}p{0.07\textwidth} X X X@{}}
\toprule
 & \normalfont\textbf{Explicit} & \normalfont\textbf{Implicit} & \normalfont\textbf{Bare} \\
\midrule
Query &
\normalfont\textit{\small Which drug is carried by TF that interacts with the basal plasma membrane and is the active ingredient in Pepto-Bismol?} &
\normalfont\textit{\small Which common over-the-counter gastrointestinal remedy used for nausea, heartburn, indigestion, upset stomach, and diarrhea is indirectly associated with the basal plasma membrane through a circulating ferric iron-binding transport glycoprotein that can also carry drugs?} &
\normalfont\textit{\small Which drug commonly used for nausea, heartburn, indigestion, upset stomach, and diarrhea is associated with the basal plasma membrane?} \\
\midrule
\multicolumn{4}{@{}l}{\normalfont\textbf{Ground-truth path}: \Bpm{} $\xrightarrow{\text{interacts with}}$ \Tf{} $\xrightarrow{\text{carrier}}$ \Bs{}} \\
\midrule
\addlinespace[2pt]
\multicolumn{4}{@{}l}{\normalfont\itshape CoT} \\
\addlinespace[2pt]
Reasoning &
\normalfont $\ldots$ \emph{Pepto-Bismol}'s active ingredient is \textbf{Bismuth subsalicylate}, carried by transferrin $\ldots$ \newline \Acorrect{Bismuth subsalicylate} &
\normalfont $\ldots$ cycles between \emph{Pepto-Bismol}, \emph{Maalox}, \emph{Mylanta}, and \emph{Tums} without committing to a final candidate $\ldots$ \newline \Anone &
\normalfont $\ldots$ candidates such as \emph{Ranitidine} and \emph{Bismuth subsalicylate} are considered for the symptom set $\ldots$ \newline \Awrong{Ranitidine} \\
\midrule
\addlinespace[2pt]
\multicolumn{4}{@{}l}{\normalfont\itshape ToG} \\
\addlinespace[2pt]
Path-finding &
\normalfont $d_1$: \Bpm{} $\to$ \emph{BEST1} \,\Xmark \newline
$d_2$: \emph{BMPR2} $\to$ \emph{Dibotermin alfa} \,\Xmark &
\normalfont $d_1$: \Bpm{} $\to$ \emph{ABCC1}/\emph{AQP1} \,\Xmark \newline
$d_2$: \emph{ABCC1} $\to$ \emph{Clofazimine} \,\Xmark &
\normalfont $d_1$: \Bpm{} $\to$ \emph{ABCC1}/\emph{AQP5} \,\Xmark \newline
$d_2$: \emph{AQP5} $\to$ \emph{palmoplantar keratoderma} \,\Xmark \\
Reasoning &
\normalfont $\ldots$ traversed triplets do not connect to a Pepto-Bismol ingredient, but the answer is recalled from external knowledge as \textbf{Bismuth subsalicylate} $\ldots$ \newline \Acorrect{Bismuth subsalicylate} &
\normalfont $\ldots$ \emph{ABCC1}/\emph{AQP1} routes do not surface a gastrointestinal remedy $\ldots$ \newline \Awrong{(no GI remedy identified)} &
\normalfont $\ldots$ no path connects \Bpm{} to a GI remedy; candidate-class reasoning lands elsewhere $\ldots$ \newline \Awrong{Ranitidine} \\
\midrule
\addlinespace[2pt]
\multicolumn{4}{@{}l}{\normalfont\itshape \sys{}} \\
\addlinespace[2pt]
Path-finding &
\normalfont $d_1$: \Bpm{} $\to$ \Tf{} \,\Cmark \newline
$d_2$: \Tf{} $\to$ \Bs{} \,\Cmark &
\normalfont $d_1$: \Bpm{} $\to$ \Tf{} \,\Cmark \newline
$d_2$: \Tf{} $\to$ \Bs{} \,\Cmark &
\normalfont $d_1$: \Bpm{} $\to$ \Tf{} \,\Cmark \newline
$d_2$: \Tf{} $\to$ \Bs{} \,\Cmark \\
Reasoning &
\normalfont $\ldots$ \textbf{TF} interacts with the \textbf{basal plasma membrane} and carries \textbf{Bismuth subsalicylate}, the active ingredient in Pepto-Bismol $\ldots$ \newline \Acorrect{Bismuth subsalicylate} &
\normalfont $\ldots$ \textbf{basal plasma membrane} $\to$ \textbf{TF} $\to$ \textbf{Bismuth subsalicylate}, a common GI remedy $\ldots$ \newline \Acorrect{Bismuth subsalicylate} &
\normalfont $\ldots$ \textbf{basal plasma membrane} $\to$ \textbf{TF} $\to$ \textbf{Bismuth subsalicylate} for GI symptoms $\ldots$ \newline \Acorrect{Bismuth subsalicylate} \\
\bottomrule
\end{tabularx}
\caption{\textbf{Case study, 2-hop, data\_id=3786.}
ToG misses the ground-truth first hop from \Bpm{} to \Tf{} under all
three query formulations and lands on unrelated transporters. It still
produces the correct final answer on the explicit query by leveraging the
Pepto-Bismol cue in the question itself, but fails on implicit and bare.
\sys{} recovers the entire ground-truth path at both depths in all three
settings.}
\label{tab:appendix_case_3786}
\end{table*}

\begin{table*}[p]
\centering\footnotesize
\setlength{\tabcolsep}{6pt}
\renewcommand{\arraystretch}{1.3}
\newcommand{\Det}{\textit{Delayed eruption of teeth}}
\newcommand{\Thra}{\textit{THRA}}
\newcommand{\Dex}{\textit{Dextrothyroxine}}
\newcommand{\Hyp}{\textit{hyperlipidemia}}
\begin{tabularx}{\textwidth}{@{}>{\bfseries}p{0.07\textwidth} X X X@{}}
\toprule
 & \normalfont\textbf{Explicit} & \normalfont\textbf{Implicit} & \normalfont\textbf{Bare} \\
\midrule
Query &
\normalfont\textit{\small Delayed eruption of teeth is associated with THRA; which disease characterized by an excess of lipids in the blood is indicated for the THRA-targeting drug Dextrothyroxine?} &
\normalfont\textit{\small Delayed eruption of teeth can be associated with impaired thyroid hormone signaling involving a receptor targeted by a synthetic thyroxine analogue; this points to what metabolic disorder characterized by excess lipids in the blood?} &
\normalfont\textit{\small Delayed eruption of teeth is associated with what disease characterized by an excess of lipids in the blood?} \\
\midrule
\multicolumn{4}{@{}l}{\normalfont\textbf{Ground-truth path}: \Det{} $\xrightarrow{\text{associated with}}$ \Thra{} $\xrightarrow{\text{target}}$ \Dex{} $\xrightarrow{\text{indication}}$ \Hyp{}} \\
\midrule
\addlinespace[2pt]
\multicolumn{4}{@{}l}{\normalfont\itshape CoT} \\
\addlinespace[2pt]
Reasoning &
\normalfont $\ldots$ \textbf{THRA} regulates thyroid signaling; the THRA-targeting drug used for blood lipid disorders is \textbf{Dextrothyroxine}, indicated for \textbf{hyperlipidemia} $\ldots$ \newline \Acorrect{hyperlipidemia} &
\normalfont $\ldots$ thyroid hormone receptor reasoning meanders across multiple disorders without settling on a definite answer $\ldots$ \newline \Anone &
\normalfont $\ldots$ candidate lipid disorders include \emph{familial hypercholesterolemia}, but the final answer settles on a different condition $\ldots$ \newline \Awrong{familial hypercholesterolemia} \\
\midrule
\addlinespace[2pt]
\multicolumn{4}{@{}l}{\normalfont\itshape ToG} \\
\addlinespace[2pt]
Path-finding &
\normalfont $d_1$: \Det{} $\to$ \Thra{} \,\Cmark \newline
$d_2$: \emph{Abnormality of dental eruption} $\to$ \emph{neighbor syndromes} \,\Xmark \newline
$d_3$: --- (path lost) &
\normalfont $d_1$: \Det{} $\to$ \Thra{} \,\Cmark \newline
$d_2$: \emph{Delayed eruption of primary teeth} $\to$ \emph{Cockayne syndrome} \,\Xmark \newline
$d_3$: --- &
\normalfont $d_1$: \Det{} $\to$ \Thra{} \,\Cmark \newline
$d_2$: \emph{Abnormality of dental eruption} $\to$ \emph{neighbor syndromes} \,\Xmark \newline
$d_3$: --- \\
Reasoning &
\normalfont $\ldots$ retrieved triplets do not connect THRA to a lipid disorder, but the THRA-Dextrothyroxine-hyperlipidemia mechanism is recalled from prior knowledge $\ldots$ \newline \Acorrect{hyperlipidemia} &
\normalfont $\ldots$ no lipid disorder surfaced in the traversed subgraph; reasoning loops between thyroid-related conditions $\ldots$ \newline \Awrong{Resistance to Thyroid Hormone} &
\normalfont $\ldots$ same as the CoT-style fallback; the answer drifts to a related lipid disorder $\ldots$ \newline \Awrong{familial hypercholesterolemia} \\
\midrule
\addlinespace[2pt]
\multicolumn{4}{@{}l}{\normalfont\itshape \sys{}} \\
\addlinespace[2pt]
Path-finding &
\normalfont $d_1$: \Det{} $\to$ \Thra{} \,\Cmark \newline
$d_2$: \Thra{} $\to$ \Dex{} \,\Cmark \newline
$d_3$: \Dex{} $\to$ \Hyp{} \,\Cmark &
\normalfont $d_1$: \Det{} $\to$ \Thra{} \,\Cmark \newline
$d_2$: \Thra{} $\to$ \Dex{} \,\Cmark \newline
$d_3$: \Dex{} $\to$ \Hyp{} \,\Cmark &
\normalfont $d_1$: \Det{} $\to$ \Thra{} \,\Cmark \newline
$d_2$: \Thra{} $\to$ \Dex{} \,\Cmark \newline
$d_3$: \Dex{} $\to$ \Hyp{} \,\Cmark \\
Reasoning &
\normalfont $\ldots$ \textbf{THRA}-targeting drug \textbf{Dextrothyroxine} is indicated for \textbf{hyperlipidemia} (and familial hyperlipidemia) $\ldots$ \newline \Acorrect{hyperlipidemia} &
\normalfont $\ldots$ the synthetic thyroxine analogue targeting \textbf{THRA} is \textbf{Dextrothyroxine}, indicated for \textbf{hyperlipidemia} $\ldots$ \newline \Acorrect{hyperlipidemia} &
\normalfont $\ldots$ \textbf{Delayed eruption of teeth} $\to$ \textbf{THRA} $\to$ \textbf{Dextrothyroxine} $\to$ \textbf{hyperlipidemia} $\ldots$ \newline \Acorrect{hyperlipidemia} \\
\bottomrule
\end{tabularx}
\caption{\textbf{Case study, 3-hop, data\_id=4416.}
ToG recovers only one of the three ground-truth triplets even on the
explicit query, landing on the THRA seed but failing to trace through
\Dex{}. The THRA to Dextrothyroxine to hyperlipidemia link is then filled
in from prior knowledge to answer the explicit query correctly, but the
implicit and bare formulations strip enough mechanism that the fallback
no longer works. \sys{} recovers all three hops under every
formulation.}
\label{tab:appendix_case_4416}
\end{table*}

\begin{table*}[p]
\centering\footnotesize
\setlength{\tabcolsep}{6pt}
\renewcommand{\arraystretch}{1.3}
\newcommand{\Nd}{\textit{Neurodevelopmental delay}}
\newcommand{\Thrb}{\textit{THRA}}
\newcommand{\Drn}{\textit{Dronedarone}}
\newcommand{\Afi}{\textit{atrial fibrillation}}
\begin{tabularx}{\textwidth}{@{}>{\bfseries}p{0.07\textwidth} X X X@{}}
\toprule
 & \normalfont\textbf{Explicit} & \normalfont\textbf{Implicit} & \normalfont\textbf{Bare} \\
\midrule
Query &
\normalfont\textit{\small Which disease, characterized by a supraventricular arrhythmia with absent consistent P waves and an irregular ventricular response, is an indication for the THRA-targeting drug Dronedarone, where THRA is associated with Neurodevelopmental delay?} &
\normalfont\textit{\small Which supraventricular arrhythmia, marked by absent organized P waves and an irregularly irregular ventricular response, could be linked to neurodevelopmental delay through disrupted thyroid hormone-regulated transcription and the target profile of a rhythm-control antiarrhythmic drug?} &
\normalfont\textit{\small Which disease marked by a rapid, irregular rhythm arising in the upper chambers of the heart is associated with Neurodevelopmental delay?} \\
\midrule
\multicolumn{4}{@{}l}{\normalfont\textbf{Ground-truth path}: \Nd{} $\xrightarrow{\text{associated with}}$ \Thrb{} $\xrightarrow{\text{target}}$ \Drn{} $\xrightarrow{\text{indication}}$ \Afi{}} \\
\midrule
\addlinespace[2pt]
\multicolumn{4}{@{}l}{\normalfont\itshape CoT} \\
\addlinespace[2pt]
Reasoning &
\normalfont $\ldots$ description matches \textbf{atrial fibrillation}; \textbf{Dronedarone} is indicated for atrial fibrillation/flutter $\ldots$ \newline \Acorrect{atrial fibrillation} &
\normalfont $\ldots$ the described rhythm-control drug acting on a thyroid receptor points to \textbf{Dronedarone}, indicated for atrial fibrillation $\ldots$ \newline \Acorrect{atrial fibrillation} &
\normalfont $\ldots$ neurodevelopmental syndromes with rhythm abnormalities are considered, with the final pick drifting away from atrial fibrillation $\ldots$ \newline \Awrong{Wolff-Parkinson-White syndrome} \\
\midrule
\addlinespace[2pt]
\multicolumn{4}{@{}l}{\normalfont\itshape ToG} \\
\addlinespace[2pt]
Path-finding &
\normalfont $d_1$: \Nd{} $\to$ \emph{Beckwith-Wiedemann syndrome} \,\Xmark \newline
$d_2$: --- (THRA branch missed) \newline
$d_3$: --- &
\normalfont $d_1$: \Nd{} $\to$ \emph{tuberous sclerosis}/\emph{CARS1} \,\Xmark \newline
$d_2$: \emph{CARS1} $\to$ \emph{Cysteine} \,\Xmark \newline
$d_3$: --- &
\normalfont $d_1$: \Nd{} $\to$ \emph{CARS1}/\emph{KCNK4} \,\Xmark \newline
$d_2$: \emph{Delayed social development} $\to$ \emph{Turner syndrome} \,\Xmark \newline
$d_3$: --- \\
Reasoning &
\normalfont $\ldots$ traversed subgraph misses the Dronedarone link; the question's description of the arrhythmia recovers \textbf{atrial fibrillation} from prior knowledge $\ldots$ \newline \Acorrect{atrial fibrillation} &
\normalfont $\ldots$ ATP-binding parent-class reasoning still surfaces \textbf{atrial fibrillation} as the matching arrhythmia $\ldots$ \newline \Acorrect{atrial fibrillation} &
\normalfont $\ldots$ Turner syndrome is associated with neurodevelopmental delay and carries cardiovascular risk $\ldots$ \newline \Awrong{Turner syndrome} \\
\midrule
\addlinespace[2pt]
\multicolumn{4}{@{}l}{\normalfont\itshape \sys{}} \\
\addlinespace[2pt]
Path-finding &
\normalfont $d_1$: \Nd{} $\to$ \Thrb{} \,\Cmark \newline
$d_2$: \Thrb{} $\to$ \Drn{} \,\Cmark \newline
$d_3$: \Drn{} $\to$ \Afi{} \,\Cmark &
\normalfont $d_1$: \Nd{} $\to$ \Thrb{} \,\Cmark \newline
$d_2$: \Thrb{} $\to$ \Drn{} \,\Cmark \newline
$d_3$: \Drn{} $\to$ \Afi{} \,\Cmark &
\normalfont $d_1$: \Nd{} $\to$ \Thrb{} \,\Cmark \newline
$d_2$: \Thrb{} $\to$ \Drn{} \,\Cmark \newline
$d_3$: \Drn{} $\to$ \Afi{} \,\Cmark \\
Reasoning &
\normalfont $\ldots$ \textbf{Dronedarone} is indicated for \textbf{atrial fibrillation} (supraventricular arrhythmia with absent P waves) $\ldots$ \newline \Acorrect{atrial fibrillation} &
\normalfont $\ldots$ \textbf{Dronedarone} targets \textbf{THRA} (linked to \textbf{neurodevelopmental delay}) and is indicated for \textbf{atrial fibrillation} $\ldots$ \newline \Acorrect{atrial fibrillation} &
\normalfont $\ldots$ \textbf{neurodevelopmental delay} $\to$ \textbf{THRA} $\to$ \textbf{Dronedarone} $\to$ \textbf{atrial fibrillation} $\ldots$ \newline \Acorrect{atrial fibrillation} \\
\bottomrule
\end{tabularx}
\caption{\textbf{Case study, 3-hop, data\_id=4423.}
ToG recovers none of the three ground-truth triplets, never reaching the
THRA seed, yet still answers the explicit and implicit queries correctly
because both spell out enough mechanism for the LLM to recall the
atrial-fibrillation indication of Dronedarone. Once the bare query
removes the drug-mechanism cue, ToG drifts to Turner syndrome. \sys{}
traces the full path and stays correct.}
\label{tab:appendix_case_4423}
\end{table*}

\section{Use of AI Assistants}
\label{sec:appendix:ai_assistants}

During manuscript preparation, the authors employed AI-based writing
assistants solely for language refinement, including grammar, phrasing,
and clarity. The research design, methodology, experiments, analyses,
and reported findings remain entirely the authors' own contributions.